\documentclass[a4paper,fleqn]{cas-dc}

\usepackage[numbers,super,sort&compress]{natbib}
\setcitestyle{comma,space}
\newcommand{\citepunc}[1]{\textsuperscript{\citealp{#1}}}

\usepackage{multirow}
\usepackage{color}
\usepackage{xcolor}
\usepackage{colortbl}
\usepackage{graphicx}
\usepackage{subcaption}
\usepackage[inline]{enumitem}
\usepackage{bm}
\usepackage{amsmath}
\usepackage{stfloats}
\usepackage{underscore}
\usepackage{longtable}
\usepackage{chngcntr}
\usepackage{makecell}

\usepackage{listings}
\usepackage{xcolor}

\usepackage[most]{tcolorbox}
\usepackage{multicol}
\usepackage{float} 

\lstdefinelanguage{prompt}{
  basicstyle=\ttfamily\small,
  breaklines=true,
  frame=single,
  backgroundcolor=\color{gray!5},
  columns=fullflexible,
  keepspaces=true,
  showstringspaces=false,
  morekeywords={DATA, FORMAT, EVALUATION, CRITERIA, OUTPUT, SCORING, RUBRIC, PATIENT, RECORD},
  keywordstyle=\bfseries\color{blue!40!black},
}

\definecolor{LightRed}{rgb}{1,0.92,0.92}
\definecolor{LightOrange}{rgb}{1,0.95,0.88}
\definecolor{LightYellow}{rgb}{1.0,1.0,0.84}
\definecolor{LightGreen}{rgb}{0.9,1.0,0.88}
\definecolor{LightCyan}{rgb}{0.9,1,1}
\definecolor{LightBlue}{rgb}{0.9,0.94,1}
\definecolor{ContextBg}{RGB}{248,232,225} 
\newtcolorbox{researchcontext}{
  colback=ContextBg,
  colframe=white, 
  boxrule=0pt,
  arc=0pt,
  left=10pt,
  right=10pt,
  top=10pt,
  bottom=10pt,
  width=\textwidth,
  enlarge left by=0mm,
  enlarge right by=0mm
}

\makeatletter

\def\ps@first{%
   \let\@oddhead\@empty
   \let\@evenhead\@empty
   \def\@oddfoot{\hfil\thepage\hfil} 
   \def\@evenfoot{\hfil\thepage\hfil}
}

\def\ps@headings{%
      \let\@oddhead\@empty
      \let\@evenhead\@empty
      \def\@oddfoot{\hfil\thepage\hfil} 
      \def\@evenfoot{\hfil\thepage\hfil}
}
\renewcommand{\printorcid}[1]{}
\makeatother

\definecolor{myrefred}{RGB}{180,0,0}
\AtBeginDocument{%
  \hypersetup{
    linkcolor={myrefred},
    filecolor=myrefred,
    menucolor=myrefred
  }%
}

\renewcommand{\ttdefault}{lmtt}

\ExplSyntaxOn
\RenewDocumentCommand \emailauthor { m m }
{
  \int_gincr:N \g_ead_int
  \seq_gput_right:Nn \g_stm_ead_seq
  {
    { \ttfamily \tl_to_str:n { #1 } }
  }
}
\ExplSyntaxOff

\begin{document}



\title [mode = title]{From Statistical Fidelity to Clinical Consistency: Scalable Generation and Auditing of Synthetic Patient Trajectories}  

\author[1]{\textcolor{black}{Guanglin} Zhou}
\ead{guanglin.zhou@uq.edu.au}
\cormark[1]

\author[2]{\textcolor{black}{Armin} Catic}
\author[3]{\textcolor{black}{Motahare} Shabestari}
\author[2]{\textcolor{black}{Matthew} Young}
\author[4]{\textcolor{black}{Chaiquan} Li}
\author[4]{\textcolor{black}{Katrina} Poppe}
\author[1,2]{\textcolor{black}{Sebastiano} Barbieri}
\ead{s.barbieri@uq.edu.au}
\cormark[1]
\cortext[1]{Corresponding author}
\affiliation[1]{organization={The University of Queensland},
  city={Brisbane},
  state={QLD},
  country={Australia}}
\affiliation[2]{organization={The University of New South Wales},
  city={Sydney},
  state={NSW},
  country={Australia}}
\affiliation[3]{organization={Shahid Sadoughi University of Medical Sciences and Health Services},
  city={Yazd},
  country={Iran}}
\affiliation[4]{organization={The University of Auckland},
  city={Auckland},
  country={New Zealand}}
\maketitle


\begin{abstract}
\textsf{\textbf{Background}}
Access to electronic health records (EHRs) for digital health research is often limited by privacy regulations and institutional barriers. 
Synthetic EHRs have been proposed as a way to enable safe and sovereign data sharing;
however, existing methods may produce records that capture overall statistical properties of real data but present inconsistencies across clinical processes and observations.

\noindent\textbf{Methods}
We developed an integrated pipeline to make synthetic patient trajectories clinically consistent through two synergistic steps: high-fidelity generation and scalable auditing.
Using the MIMIC-IV database, we trained a knowledge-grounded generative model that represents nearly 32,000 distinct clinical events, including demographics, laboratory measurements, medications, procedures, and diagnoses, while enforcing structural integrity. 
To support clinical consistency at scale, we incorporated an automated auditing module leveraging large language models to filter out clinical inconsistencies (e.g., contraindicated medications) that escape probabilistic generation.

\noindent\textbf{Findings}
We generated 18,071 synthetic patient records derived from a source cohort of 180,712 real patients. While synthetic clinical event probabilities demonstrated robust agreement (mean bias effectively 0$\cdot$00) and high correlation ($R^2$=0$\cdot$99) with the real counterparts, review of a random sample of synthetic records (N=20) by three clinicians identified inconsistencies in 45-60\% of them.
Automated auditing reduced the difference between real and synthetic data (Cohen's effect size $d$ between 0$\cdot$59 and 1$\cdot$60 before auditing, and between 0$\cdot$18 and 0$\cdot$67 after auditing).
Downstream models trained on audited data matched or even exceeded real-data performance.
We found no evidence of privacy risks, with membership inference performance indistinguishable from random guessing (F1-score=0$\cdot$51).

\noindent\textbf{Interpretation} Statistical fidelity is a necessary but potentially insufficient condition for creating useful synthetic clinical data. Safe synthetic data requires a framework-level approach combining knowledge-grounded generation to ensure statistical fidelity and automated auditing to support plausible clinical reasoning. The proposed framework allows simulating realistic, clinically consistent, and privacy-preserving patient trajectories at scale.

\noindent\textbf{Funding} None.
\end{abstract}





\section{Introduction}
\noindent Hospital patient care is characterized by a sequence of linked clinical events. An abnormal laboratory result may prompt further tests, lead to diagnoses, and result in medication prescriptions or procedural interventions.\citepunc{hayrinen2008definition,johnson2023mimic} 
Understanding these longitudinal care pathways is central to precision medicine and the development of reliable clinical decision support systems. 
However, access to real-world electronic health record (EHR) data for research purposes remains severely limited by privacy regulations, ethical requirements, and institutional silos, hindering reproducibility and methodological progress in digital health.\citepunc{bani2020privacy,keshta2021security}

Synthetic EHR data generation offers a privacy-preserving solution to enable sovereign data sharing across borders and institutions, particularly for underserved populations.\citepunc{gonzales2023synthetic,mcduff2023synthetic,van2024synthetic,ghosheh2024survey} 
Beyond data sharing, synthetic cohorts are increasingly vital for simulating future patient trajectories (e.g., in digital twins\citepunc{katsoulakis2024digital}), and validating fair clinical artificial intelligence tools against local demographics.\citepunc{liu2025conditional,Marchesi2025-lv} 
Yet, for these downstream applications to be reliable, synthetic data must preserve clinically relevant temporal and causal relationships among heterogeneous clinical events. 
These relationships reflect how clinical decisions are made in practice and are essential for synthetic data to be clinically consistent.\citepunc{manktelow2022clinical}

Despite rapid progress, the faithful representation of patient trajectories in synthetic EHR data remains challenging.
Early deep learning approaches, including Generative Adversarial Networks or Variational Auto-Encoders, often relied on coarse aggregation or narrow subsets of clinical events, failing to capture the full complexity of longitudinal care.\citepunc{Zhou2025-an,baowaly2019synthesizing,rashidian2020smooth,torfi2020corgan,biswal2021eva,zhang2021synteg,yoon2023ehr}
Recent Transformer-based models improve long-range dependencies, but typically restrict the scope of clinical data or use tokenization strategies prone to producing non-existent medical concepts.\citepunc{vaswani2017attention,theodorou2023synthesize,Zhou2025-mt,renc2024zero,waxler2025generative}
Consequently, existing generative models struggle to represent the full spectrum of patient care and lack basic structural integrity.

Furthermore, resolving structural issues reveals a more subtle challenge: statistical fidelity does not guarantee clinical consistency. 
Even when generating valid clinical concepts, probabilistic models inevitably produce semantic contradictions, such as pregnancy-related procedures being associated with a male patient or medications being prescribed without a plausible clinical indication. 
These errors are not necessarily failures in terms of event generation, but failures to reflect real-world clinical reasoning. 
Identifying them creates a scalability paradox: review by expert clinicians is the gold standard but is time-consuming, subjective, and impractical for large synthetic cohorts. Thus, synthetic datasets remain statistically plausible but clinically untrustworthy.

\begin{figure*}[t]
\begin{researchcontext}
\textbf{\color{red!60!black} Research in context}

\begin{multicols}{2}

\textbf{Evidence before this study}

We searched Google Scholar and PubMed for published studies on synthetic electronic health records and generative modeling of patient trajectories published between Jan 1, 2021, and Jan 1, 2026.
We used combinations of the terms (``synthetic EHR'' OR ``synthetic health records'' OR ``patient trajectories'') AND (``generative model'' OR ``Transformer'' OR ``deep learning'') AND (``longitudinal'' OR ``temporal''). 
We also reviewed reference lists of relevant articles. 
Prior research has demonstrated that deep learning approaches, particularly Transformer-based models, can generate EHR data with good statistical fidelity. However, prior studies generally present two important limitations: (1) the generative models focused on limited subsets of medical events, relied on coarse or fragmented representations, leading to non-existent or structurally invalid medical concepts; and (2) validation is primarily conducted using aggregate statistical metrics. 
We found little evidence of approaches that both generate complete, realistic patient trajectories, and validate and audit the clinical consistency before downstream use.

\par\medskip
\textbf{Added value of this study}

To our knowledge, this is the first study to show that clinically consistent synthetic EHR data requires both realistic generation and scalable automated auditing, rather than focusing only on statistical fidelity.
We introduce Coogee, a framework that overcomes key limitations of prior models by representing nearly 32,000 heterogeneous clinical events---including demographics, laboratories, medications, and procedures---using biomedical knowledge, thereby preventing the generation of non-existent medical concepts. 
We further address a central challenge in synthetic data validation: while statistical fidelity is necessary, it does not ensure the validity of represented clinical processes, and manual review by clinicians is inconsistent and cannot scale to large synthetic datasets. 
By integrating an automated Large Language Model (LLM) auditor, our framework identifies and filters out clinically invalid records (e.g., a male with pregnancy-related procedures) or violations of basic clinical reasoning. We show that clinically valid synthetic cohorts emerge only when realistic generation is paired with rigorous, scalable auditing.

\par\medskip
\textbf{Implications of all the available evidence}

Our findings suggest that synthetic EHR data should no longer be evaluated or deployed solely on the basis of overall statistical similarity to real-world data. 
The observed gap between statistical fidelity and clinical consistency introduces risks for research and simulation, particularly when synthetic data are used without explicit clinical review. Scalable, automated auditing should therefore be adopted as a standard component of synthetic data pipelines to ensure appropriate governance and trust. This framework enables the safe and reproducible use of synthetic patient data for the development and evaluation of clinical artificial intelligence applications in settings where access to real-world EHRs is restricted.

\end{multicols}
\end{researchcontext}
\end{figure*}

To close these gaps, we introduce Coogee, a two-step framework for generating and auditing clinically consistent synthetic patient trajectories. 
First, to ensure structural integrity and full-spectrum coverage, the generative component models nearly 32,000 distinct clinical events by grounding representations in biomedical knowledge rather than arbitrary tokenization. 
Second, to ensure clinical consistency at scale, Coogee leverages large language models (LLMs) for automated auditing. 
Because modern LLMs possess sophisticated clinical reasoning capabilities, including near-expert USMLE performance, their evaluations strongly correlate with human clinicians.\citepunc{singhal2023large,hager2024evaluation}
This module detects and filters out the semantic inconsistencies that inevitably escape probabilistic generation, providing a practical alternative to review by human clinicians.
We evaluate Coogee using the Medical Information Mart for Intensive Care (MIMIC-IV) dataset.\citepunc{johnson2023mimic} We show that realistic, clinically consistent, and privacy-preserving synthetic cohorts emerge only when realistic generation is paired with rigorous automated timeline auditing.

\section{Methods}

\subsection{Data sources and study cohort}
\noindent 
We utilized the Medical Information Mart for Intensive Care (MIMIC-IV) database (v2.2), a comprehensive, de-identified EHR dataset sourced from Beth Israel Deaconess Medical Center (Boston, MA, USA).\citepunc{johnson2023mimic}
The study cohort included all patients with at least one recorded hospital admission.
Admissions were ordered chronologically by admission time to preserve the temporal sequence of encounters. Clinical concepts were aggregated within each admission, forming structured longitudinal records. 
The final cohort (Table~\ref{816874121706}) comprised 180,712 unique patients, partitioned at the patient level into training (81\%, N=146,377), validation (9\%, N=16,264), and test (10\%, N=18,071) sets to prevent data leakage. 
The training set includes approximately 121 million clinical events, providing a large-scale foundation for generative modeling (Appendix~\ref{796598494627}).

\subsection{The generative framework}
\label{947059557677}
\noindent Existing transformer-based generative models often prioritize efficiency over structural integrity, utilizing sub-word tokenization strategies that fragment medical codes (e.g., splitting the diagnostic code \texttt{E11.9} into `\texttt{E11}' and `\texttt{.9}').\citepunc{renc2024zero,waxler2025generative}
This introduces a critical safety risk: the recombination of fragments into non-existent hallucinated codes. 
To eliminate this failure mode, we implemented an \emph{atomic tokenization strategy}, representing every distinct clinical concept as a single, indivisible token. 
This ensures a strict one-to-one mapping to recognized ontology concepts (Table~\ref{828348301138}). We mitigated the computational cost of this expanded vocabulary via efficient matrix factorization of the input embedding layer (Appendix~\ref{045304954792}).

To capture the irregular cadence of disease progression, we explicitly encoded temporal information using discrete \emph{time-gap tokens} (spanning minutes to months), enabling the framework to faithfully represent the variable tempos of care.\citepunc{theodorou2023synthesize,Zhou2025-mt}
Continuous laboratory values were discretised into decile-based quantile bins, and uniformly sampled within that range during sequence reconstruction to retain numerical realism.
Finally, to improve representation of rare diseases and complex comorbidities, we anchored each clinical token in biomedical knowledge.
Using the PrimeKG knowledge graph, we projected structured biomedical relationships (e.g., pathways linking a drug to a disease) and semantic definitions into the model's latent space.\citepunc{chandak2023building} These auxiliary signals enable the model to generalize based on medical semantics rather than isolated frequency statistics ( Appendix~\ref{206223520903}).

\subsection{Scalable automated auditing}
\noindent Although our generative framework enforces structural validity, probabilistic generation remains prone to \emph{semantic hallucinations}. As a result, synthetic trajectories may be syntactically correct yet clinically inconsistent (e.g., conflicting demographics or broken causal chains). Because manual review by clinicians is subjective and infeasible for large cohorts, we developed a scalable automated auditing module leveraging  the clinical reasoning capabilities of large language models (LLMs).\citepunc{singhal2023large}

Using a local instance of the open-weights Qwen3-30B model,\footnote{\url{https://huggingface.co/Qwen/Qwen3-30B-A3B-Instruct-2507}} we prompted the LLM to act in the role of a medical expert and evaluate complete patient trajectories against three dimensions of clinical consistency: (1) \emph{demographic alignment} verifying consistency between biological attributes (such as age and sex) and clinical events (e.g., verifying the plausibility of sex-specific diagnoses); (2) \emph{clinical reasoning} ensuring adherence to necessary causal chains, such as medications being supported by relevant diagnoses or laboratory findings; (3) \emph{temporal plausibility} confirming physically and clinically feasible event sequences. The LLM scored each record on a 10-point realism scale (1--2: clearly artificial; 3--4: largely synthetic; 5--6: mixed realism; 7--8: mostly realistic; 9--10: indistinguishable from real-world EHRs) and only trajectories scoring at least 7/10 (indicating ``Mostly Realistic'' to ``Indistinguishable'') were retained in the final synthetic dataset.

To validate the reliability of automated auditing, we compared the LLM auditor against three blinded clinicians and alternative state-of-the-art LLMs (GPT-5, Gemini-3-Pro, Qwen-3-Max), on a random sample of 40 complete patient trajectories (20 real MIMIC-IV test records and 20 Coogee synthetic records).
Full details of the auditing prompt and scoring rubric are provided in the Supplementary Appendix~\ref{941068451355}.

\subsection{Experimental protocol and evaluation metrics}
\noindent We trained the generative model via causal language modeling. Cohort generation was initialized with held-out test population demographics, followed by autonomous sampling of clinical events and temporal gaps.
We applied constrained decoding to enforce structural integrity (e.g., restricting laboratory test tokens to valid quantile bins) and utilized nucleus sampling ($\text{top-p}=0.98$, $\text{temperature}=1.0$) until an \texttt{END\_RECORD} or \texttt{DEATH} token.
Finally, the automated auditing module filtered the raw synthetic trajectories, retaining only those meeting rigorous consistency thresholds for downstream utility
(Supplementary Appendix~\ref{617393837400}).

We evaluated synthetic EHR quality along four dimensions: statistical fidelity, clinical consistency, downstream utility, and privacy preservation.
Statistical fidelity was assessed via Bland-Altman analyses and the coefficient of determination ($R^2$) to confirm the agreement and the correlation in code prevalence and co-occurrence. Structural and temporal properties, including visit density and length-of-stay, were compared using the Kolmogorov–Smirnov statistic and overlap coefficients.
Clinical consistency was evaluated through blinded clinician review and automated auditing, using inter-rater reliability for alignment and Cohen's $d$ to quantify distinguishability.
Downstream utility was measured via a Train-on-Synthetic, Test-on-Real (TSTR) protocol predicting four clinical outcomes (phenotyping, in-hospital mortality, length-of-stay, 30-day readmission), evaluated using AUROC and F1-score.
Finally, privacy preservation was assessed via Membership Inference Attacks (MIA; evaluated against a random guessing baseline F1=0$\cdot$5) and Attribute Inference Attacks (AIA).

\subsection{Role of the funding source}
\noindent The institutions employing the authors of this study had no role in study design, data collection, data analysis, data interpretation, or writing of the report.

\section{Results}
\subsection{Statistical fidelity of the generative framework}

\begin{figure*}[!htb]
\centering
\includegraphics[width=0.96\linewidth,keepaspectratio]{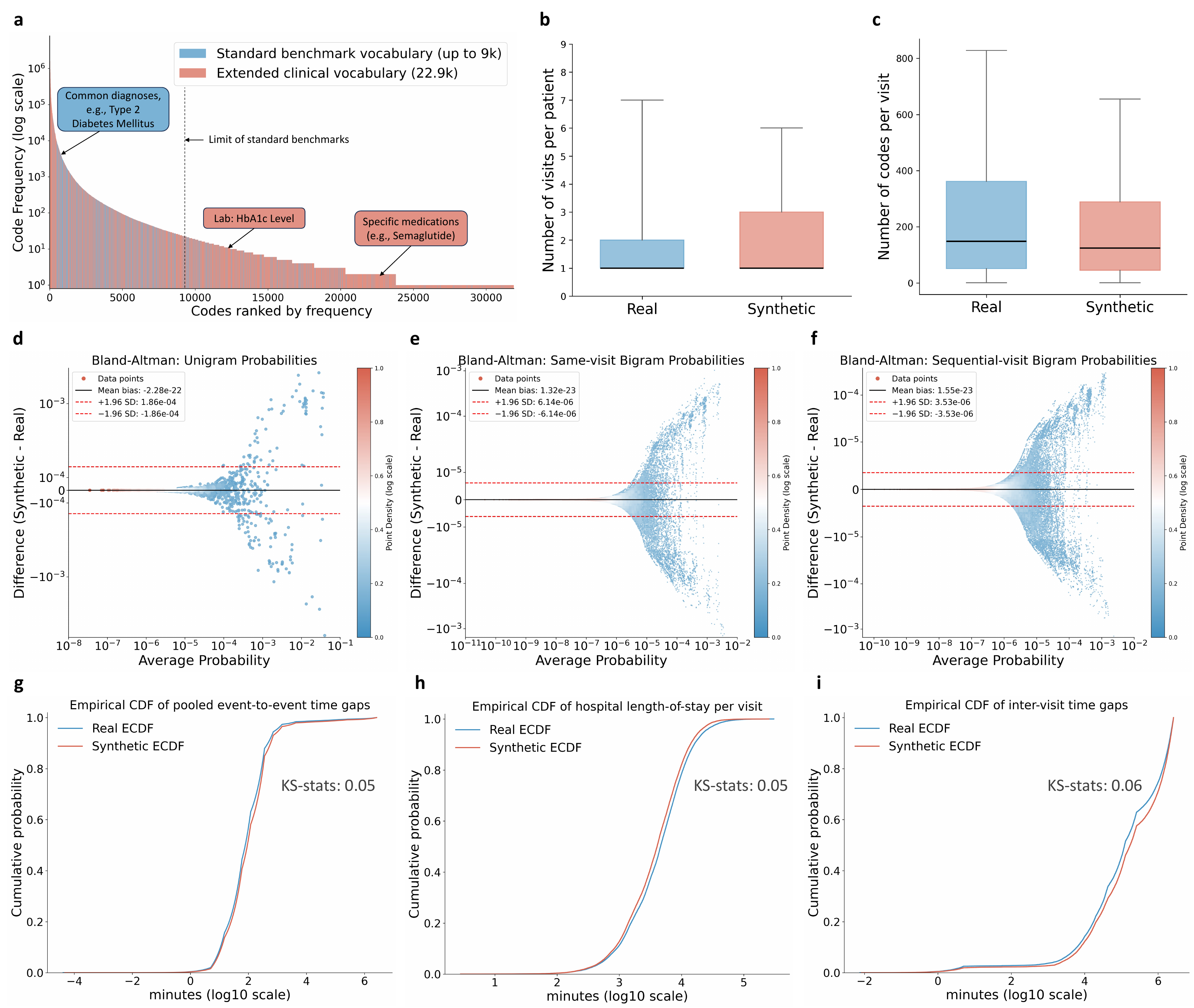}
\caption{\textbf{Fidelity and structural evaluation of synthetic EHR data.} 
\textbf{a} Vocabulary coverage across 31,901 concepts. While standard benchmarks (blue) capture high-frequency diagnoses, Coogee models the ``long tail'' of clinical care (red), including specific medications (e.g., Semaglutide) and laboratory tests (e.g., HbA1c) required for high-fidelity disease modeling.
\textbf{b-c}, Structural characteristics comparing real (blue) and synthetic (red) cohorts. The synthetic data mirrors real-world distributions for visits per patient and codes per visit, with minor truncation in the extreme upper tail due to model context window limits.
\textbf{d-f}, Probabilistic agreement assessed via density-scaled Bland-Altman plots. Minimal mean bias (0·00) and narrow 95\% limits of agreement across unigrams (marginal code probabilities), same-visit co-occurrences, and sequential-visit dependencies confirm that the model reproduces phenotype prevalence and care pathways without systematic bias.
\textbf{g-i}, Temporal fidelity evaluation using empirical cumulative distribution functions (ECDFs). Distributional alignment (KS-stats $\leq$0$\cdot$06) across event gaps, length of stay, and inter-visit intervals confirms the model's ability to capture the irregular cadence of healthcare. 
}
\label{fidelity_ba}
\end{figure*}

\begin{figure*}[!tbh]
\centering
\includegraphics[width=1.0\linewidth,keepaspectratio]{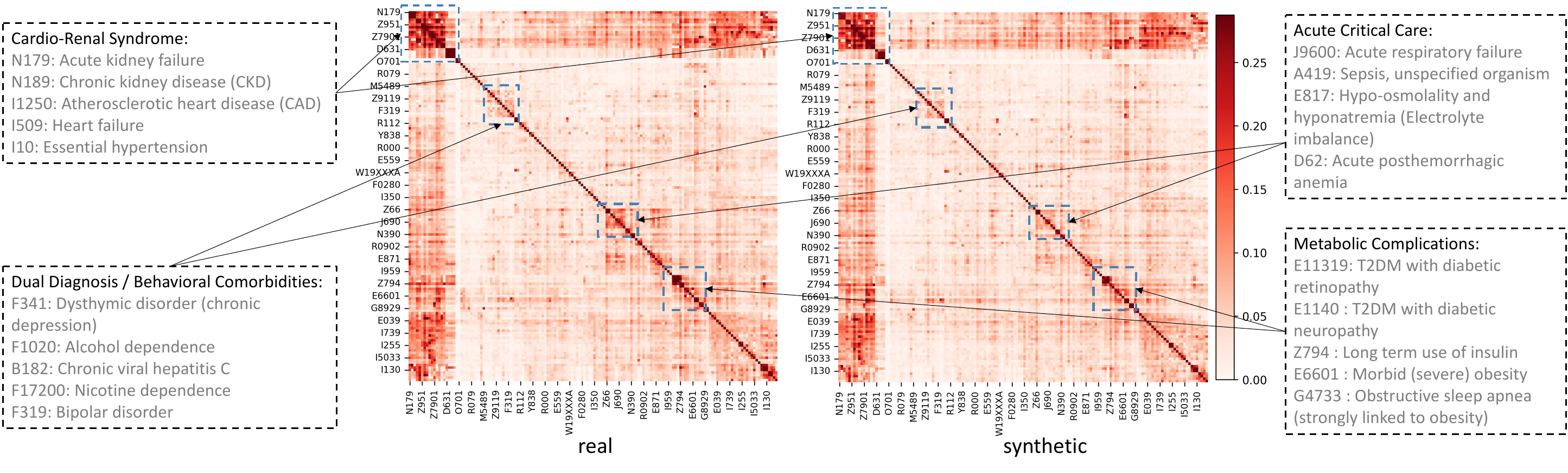}
\caption{
\textbf{Preservation of complex clinical phenotypes and syndromic co-occurrences.} 
Co-occurrence heatmaps of the top 150 diagnosis codes in Real (left) and Synthetic (right) cohorts, with axes sorted via hierarchical clustering. High Pearson ($r=0$$\cdot$$93$) and Spearman ($\rho=0$$\cdot$$88$) correlations demonstrate that Coogee generates coherent clinical syndromes rather than isolated codes. Annotated regions highlight the faithful reproduction of higher-order disease mechanisms, including the Cardio-Renal Syndrome and metabolic disease progression.
}
\label{heatmap_figure}
\end{figure*}

\noindent We first assessed whether Coogee preserves the statistical and structural properties of real EHR data (Figures~\ref{fidelity_ba}-\ref{heatmap_figure}).
We compared the synthetic cohort against the full real test set, excluding incomplete synthetic trajectories truncated by the 2,048-token context window to ensure semantically complete patient histories.

As shown in Figure~\ref{fidelity_ba}a, clinical data follows a heavy-tailed Zipfian distribution. While prior benchmarks typically restrict modeling to the top $\sim$9,000 codes, Coogee successfully models almost 32,000 distinct concepts, including $\sim$22.8k ``long-tail'' codes.\citepunc{theodorou2023synthesize}
In terms of trajectory length (Figure~\ref{fidelity_ba}b), the synthetic cohort had a median of 1$\cdot$0 visits (interquartile interval [IQI], 1$\cdot$0-3$\cdot$0), closely matching the real cohort median of 1$\cdot$0 (IQI, 1$\cdot$0-2$\cdot$0; overlap coefficient 0$\cdot$046). 
Visit density (Figure~\ref{fidelity_ba}c) similarly tracked real data with a synthetic median of 124 codes (IQI, 45–289) versus a real median of 148 (IQI, 51-362). The divergence in the mean values (225 synthetic vs. 331 real) was driven by the 99th percentile (1431 vs. 2989 codes), an expected consequence of the fixed context window (2,048 tokens in our setting). A sensitivity analysis is provided in Supplementary Appendix~\ref{635022531020}. For over 95\% of records, the distributions were well aligned (Kolmogorov-Smirnov [KS] statistic: 0$\cdot$06).\citepunc{smirnov1948table}

To evaluate probabilistic agreement between real and synthetic data we assessed the agreement between marginal code probabilities using Bland-Altman analyses~\citepunc{bland1986statistical} and correlation coefficients (Figure~\ref{correlation}).
Robust agreement was observed with an effective mean bias of 0$\cdot$00 across all granularities. Narrow 95\% limits of agreement were observed for individual clinical concepts ($-1\cdot86 \times 10^{-4}$ to $1\cdot86 \times 10^{-4}$), same-visit co-occurrences ($-6\cdot14 \times 10^{-6}$ to $6\cdot14 \times 10^{-6}$), and longitudinal sequential visits ($-3\cdot53 \times 10^{-6}$ to $3\cdot53 \times 10^{-6}$). These minimal differences confirm the model accurately reproduces phenotype prevalence and care pathway dependencies without systematic bias.

The framework accurately reflected real-world pacing across three key metrics (Figure~\ref{fidelity_ba}g–i):
\begin{enumerate*}[label=(\roman*)]
\item \emph{Pooled event-to-event gaps}: Median interval was 92 minutes (IQI, 34-271) in synthetic data vs. 78 minutes (IQI, 30-235) in real data (KS statistic=0$\cdot$05).
\item \emph{Length of stay (LOS)}: Synthetic records yielded a median LOS of 2$\cdot$81 days (IQI, 1$\cdot$19-5$\cdot$56) versus 3$\cdot$18 days (IQI, 1$\cdot$36-6$\cdot$22) in the real cohort (KS= 0$\cdot$05).
\item \emph{Inter-visit gaps}: Real patients had a median gap of 84$\cdot$6 days (IQI, 17$\cdot$7-713$\cdot$4), while synthetic patients had a median of 121$\cdot$7 days (IQI 22$\cdot$7-870$\cdot$8; KS=0$\cdot$06).
\end{enumerate*}

Hierarchical clustering to diagnosis co-occurrences  (Figure~\ref{heatmap_figure}) confirmed that Coogee generates coherent latent clinical structures and faithfully reproduces the distinct block-diagonal patterns observed in real data (Pearson correlation $r=0$$\cdot$$93$; Spearman rank correlation $\rho=0$$\cdot$$88$). The model preserved higher-order disease mechanisms, including \emph{The Cardio-Renal Syndrome} (correlating hypertension (I10), Chronic Kidney Disease (N18), and heart failure (I50)), and \emph{Metabolic Complications} (linking Type 2 Diabetes with Retinopathy (\texttt{E11.3}), Insulin use (\texttt{Z79.4}) and Morbid Obesity (\texttt{E66.0})).

\begin{table*}[!t]
\centering
\caption{\textbf{Impact of automated auditing on clinical consistency and realism.} 
Comparison of realism scores assigned by clinicians and LLMs before (Round 1) and after (Round 2) automated auditing. The Effect Size (Cohen's $d$) quantifies the statistical separation between real and synthetic records (lower indicates higher indistinguishability).
In Round 1, Rule-checking evaluators easily distinguished synthetic records ($d > 1\cdot0$, $p \le 0\cdot001$).
In Round 2, discrimination ability decreased significantly.}
\label{313819560947}
\resizebox{\textwidth}{!}{%
\begin{tabular}{ll|cccc|cccc}
\toprule
& & \multicolumn{4}{c}{\textbf{Round 1 (Unaudited)}} & \multicolumn{4}{c}{\textbf{Round 2 (Audited)}} \\
\cmidrule{3-6} \cmidrule{7-10}
\textbf{Evaluator} & \textbf{Review Strategy} & \textbf{Real} & \textbf{Syn.} & \textbf{$P$-val} & \textbf{Effect Size ($d$)} & \textbf{Real} & \textbf{Syn.} & \textbf{$P$-val} & \textbf{Effect Size ($d$)} \\
\multicolumn{2}{c|}{} & \textit{Mean (SD)} & \textit{Mean (SD)} & & & \textit{Mean (SD)} & \textit{Mean (SD)} & & \\
\midrule
\multicolumn{10}{l}{\textit{Human review}} \\
Reviewer 1 & Narrative & 7$\cdot$45 (1$\cdot$83) & 6$\cdot$15 (2$\cdot$41) & 0$\cdot$07 & 0$\cdot$59 & 6$\cdot$45 (2$\cdot$18) & 6$\cdot$00 (2$\cdot$59) & 0$\cdot$57 & 0$\cdot$18 \\
Reviewer 2 & Narrative & 6$\cdot$30 (2$\cdot$65) & 4$\cdot$80 (2$\cdot$69) & 0$\cdot$09 & 0$\cdot$55 & 5$\cdot$50 (2$\cdot$46) & 4$\cdot$05 (2$\cdot$33) & 0$\cdot$07 & 0$\cdot$59 \\
Reviewer 3 & {Rule Check} & 8$\cdot$10 (0$\cdot$99) & 5$\cdot$60 (1$\cdot$91) & {$<$0$\cdot$001} & {1$\cdot$60} & 6$\cdot$90 (1$\cdot$26) & 5$\cdot$80 (1$\cdot$89) & {0$\cdot$04} & {0$\cdot$67} \\
\midrule
\multicolumn{10}{l}{\textit{Automated review (LLMs)}} \\
GPT-5 & Logic Auditor & 6$\cdot$25 (0$\cdot$99) & 4$\cdot$80 (1$\cdot$47) & 0$\cdot$001 & 1$\cdot$13 & 6$\cdot$10 (1$\cdot$34) & 5$\cdot$15 (1$\cdot$28) & 0$\cdot$03 & 0$\cdot$71 \\
Gemini-3-Pro & Logic Auditor & 8$\cdot$25 (1$\cdot$70) & 4$\cdot$85 (2$\cdot$63) & $<$0$\cdot$001 & 1$\cdot$50 & 8$\cdot$05 (2$\cdot$20) & 5$\cdot$75 (2$\cdot$83) & 0$\cdot$008 & 0$\cdot$88 \\
Qwen-3-Max & Logic Auditor & 7$\cdot$50 (1$\cdot$86) & 5$\cdot$30 (2$\cdot$00) & {0$\cdot$001} & {1$\cdot$11} & 7$\cdot$15 (2$\cdot$20) & 6$\cdot$70 (2$\cdot$17) & 0$\cdot$53 & {0$\cdot$20} \\
\bottomrule
\end{tabular}%
}
\end{table*}

\subsection{The consistency gap: clinician and LLM-based review}
\label{944464728637}
\noindent To evaluate clinical consistency, three clinicians conducted a blinded expert review. 
The clinicians devoted considerable time (5--30 hours) to this task and review strategies differed between clinicians (Table~\ref{313819560947}, Round 1).
Reviewers 1 and 2 primarily assessed longitudinal plausibility, focusing on whether diagnoses, lab tests, and treatments formed a coherent story. The reviewers noted the presence of irrelevant and repetitive patterns in synthetic records; for example, ``major depressive disorder'' and ``nicotine dependence'' were repeated many times across records, which seems unlikely in real EHRs.
Their realism scores for real vs. synthetic records showed considerable overlap (Figure~\ref{realism-scores-round-1}a) and were not statistically significant ($p$=0$\cdot$07 and $p$=0$\cdot$09, respectively, using Wilcoxon rank-sum test).
Reviewer 3 used a ``rule-checking'' strategy, explicitly validating that metabolic syndrome patterns matched comorbidities and lab values, and that demographic variables were aligned. For example, this reviewer evaluated whether the patients' anion gap fell within plausible ranges. Under this paradigm, synthetic records were clearly distinguishable (p-value$\leq$0$\cdot$001). 
The rule-checking Reviewer 3 assessed 100\% (20/20) of real trajectories as plausible (realism score $\ge$ 6) but only 40\% (8/20) of the synthetic cohort, effectively rejecting the majority due to logical violations.

These findings expose a ``consistency gap'' where high statistical fidelity masks implausible clinical relationships.
To automate this validation, we tasked three models (GPT-5, Gemini-3-Pro, Qwen-3-Max) with the same evaluation (Appendix~\ref{941068451355}). 
The three LLMs were able to distinguish between real and synthetic records ($p\leq$0$\cdot$001), with less than 45\% of the synthetic records rated as realistic. 
Intraclass Correlation Coefficients (ICCs) indicated high agreement between the LLMs and Reviewer 3 (ICCs of 0$\cdot$34, 0$\cdot$56 and 0$\cdot$61, respectively; Figure~\ref{realism-scores-round-1}b), suggesting that LLMs can act as scalable proxies for evaluating clinical consistency.

\subsection{Impact of scalable auditing}
\noindent A second blinded evaluation was conducted using synthetic records that passed the Qwen-30B auditor filter (Score $\ge$ 7). In Round 2, discrimination ability significantly decreased across clinical reviewers (Table~\ref{313819560947}, Round 2). 
For the LLMs, which were previously the most sensitive to synthetic records, the ability to distinguish between real and synthetic data was considerably reduced. 
This reduction was most pronounced for Qwen-3-Max, where the effect size collapsed from 1$\cdot$11 to 0$\cdot$20 ($p=0\!\cdot\!53$). 
While this near-perfect indistinguishability may be partially attributable to architectural alignment with the Qwen-30B auditor, substantial improvements were also observed for unrelated models (e.g., GPT-5 effect size decreased from 1$\cdot$13 to 0$\cdot$71). 
This confirms that the auditing module addresses fundamental logical inconsistencies rather than merely overfitting to a specific model family.
Similarly, the human rule-checking expert (Reviewer 3) found the audited records more difficult to separate. The mean realism score for synthetic records rose from 5$\cdot$60 to 5$\cdot$80, and the likely-real rate improved from 40\% to 65\%, narrowing the gap to real records by more than half.

\begin{table*}[!t]
\centering
\caption{\textbf{Downstream utility and privacy preservation.} 
\textbf{Panel A:} compares the predictive performance of models trained on Real vs. Synthetic data across four clinical tasks. 
\textbf{Panel B:} quantifies privacy risks.}
\label{tab_utility_privacy}

\resizebox{\textwidth}{!}{%
\begin{tabular}{l cccccc}
\toprule
\multicolumn{7}{l}{PANEL A: downstream predictive utility} \\
\midrule
& \multicolumn{2}{c}{\textbf{AUROC (95\% CI)}} & \multicolumn{2}{c}{\textbf{F1-Score}} & \multicolumn{2}{c}{\textbf{Recall (Sensitivity)}} \\
\cmidrule(lr){2-3} \cmidrule(lr){4-5} \cmidrule(lr){6-7}
Clinical Task & Real & Synthetic & Real & Synthetic & Real & Synthetic \\
\midrule
1. Phenotyping (25-label)   & 0$\cdot$96 (0$\cdot$95--0$\cdot$96) & 0$\cdot$91 (0$\cdot$89--0$\cdot$94) & 0$\cdot$93 & 0$\cdot$90 & 0$\cdot$95 & 0$\cdot$89 \\
2. Mortality (Binary)       & 0$\cdot$76 (0$\cdot$73--0$\cdot$78) & 0$\cdot$69 (0$\cdot$65--0$\cdot$72) & 0$\cdot$72 & 0$\cdot$70 & 0$\cdot$75 & 0$\cdot$98 \\
3. Length-of-Stay (10-class)& 0$\cdot$77 (0$\cdot$76--0$\cdot$79) & 0$\cdot$74 (0$\cdot$73--0$\cdot$76) & 0$\cdot$32 & 0$\cdot$21 & 0$\cdot$34 & 0$\cdot$25 \\
4. Readmission (Binary)     & 0$\cdot$61 (0$\cdot$57--0$\cdot$64) & 0$\cdot$62 (0$\cdot$58--0$\cdot$65) & 0$\cdot$51 & 0$\cdot$62 & 0$\cdot$44 & 0$\cdot$66 \\
\bottomrule
\end{tabular}%
}

\vspace{2pt} 

\resizebox{\textwidth}{!}{%
\begin{tabular}{l l c c l}
\multicolumn{5}{l}{PANEL B: privacy preservation} \\
\midrule
Attack Scenario & {Metric} & {Baseline / Risk Limit} & {Observed Value} & {Interpretation} \\
\midrule
\multirow{2}{*}{Membership Inference (MIA)} 
  & Accuracy & 0$\cdot$50 (Random Guess) & 0$\cdot$51 & No Membership Leakage \\
  & F1-Score & 0$\cdot$50 (Random Guess) & 0$\cdot$51 & Indistinguishable \\
\addlinespace[3pt]
Attribute Inference (AIA) 
  & F1-Score & 0$\cdot$19 (Real Data Risk) & 0$\cdot$16 & Reduced Attribute Risk \\
\bottomrule
\end{tabular}%
}
\end{table*}

\subsection{Downstream utility and privacy}
\noindent Using the \emph{Train-on-Synthetic, Test-on-Real (TSTR)} protocol, we evaluated synthetic records across four diverse clinical tasks.\citepunc{esteban2017real,Zhou2025-mt}
Synthetic cohorts maintained a high degree of predictive utility, achieving performance parity or near-parity with real data baselines (Table~\ref{tab_utility_privacy}, Panel A).
In phenotype prediction, the synthetic model retained 95\% of real data discriminatory power (AUROC 0$\cdot$91 vs 0$\cdot$96).
In readmission prediction, the model achieved statistical equivalence (AUROC 0$\cdot$62 vs 0$\cdot$61) and superior sensitivity (Recall 0$\cdot$66 vs 0$\cdot$44).
For in-hospital mortality, while the synthetic model showed a reduction in overall discrimination (AUROC 0$\cdot$69 vs 0$\cdot$76), it exhibited exceptionally high sensitivity (Recall 0$\cdot$98 vs 0$\cdot$75).
A moderate drop in LOS prediction (AUROC 0$\cdot$74 vs 0$\cdot$77) reflected the removal of extreme outliers by the auditor.

We found no evidence of privacy risks (Table~\ref{tab_utility_privacy}, Panel B).
Membership Inference Attack (MIA) accuracy and F1 scores were 0$\cdot$51, equivalent to random guessing. This confirms that the generative process does not leak discernible signals regarding training set membership.
Attribute Inference Attack (AIA) risks were lower for synthetic data (F1 0$\cdot$16) than for real data (F1 0$\cdot$19), demonstrating that Coogee effectively decouples patient identity from clinical patterns.

\section{Discussion}
\noindent We presented Coogee, an integrated framework that prioritizes clinical consistency alongside statistical fidelity. Unlike previous generative architectures,\citepunc{theodorou2023synthesize,Zhou2025-mt} Coogee couples knowledge-grounded generation with scalable automated auditing to resolve the tension between statistical probability and clinical validity: the generative model learns the complex, high-dimensional distributions of patient care, while the LLM-based auditor enforces plausible clinical reasoning. This dual approach allows the system to scale to large populations without the ``hallucinations'' that often limit transformer-based health models.

Coogee generates high-fidelity synthetic cohorts that mirror real-world  complexity (e.g., multimorbidity, irregular temporal gaps) without exposing sensitive identifiers. Because the framework simulates full-spectrum individual timelines—from demographics to detailed laboratory values—it also establishes a methodological foundation for the broader vision of digital twins in healthcare.\citepunc{katsoulakis2024digital} 
While prior works have highlighted the potential of such simulations for causal inference and trial emulation, they have also cautioned against the risks of semantic inconsistencies.\citepunc{waxler2025generative} 
Our findings confirm that these risks are real: without auditing, even high-fidelity models produce trajectories that are statistically probable but clinically impossible (e.g., broken clinical pathways). By integrating the auditing module, Coogee transforms these simulations from merely ``plausible'' to ``logically sound,'' thereby increasing confidence in their use for hypothesis generation and methodological development.

Our results illustrate a fundamental discrepancy in generative AI for healthcare: the gap between \emph{statistical likelihood} and \emph{plausible clinical reasoning}. As noted in large-scale modeling efforts,\citepunc{singhal2023large} generative models prioritize the probability of the next token based on frequency, whereas clinical practice is governed by causal rules and pathophysiological constraints. This discrepancy produces a scenario where a synthetic record may appear plausible to a narrative-focused reviewer but fails a rules-based assessment, a phenomenon we term the ``clinical consistency gap.'' In our study, human reviewers focusing on narrative flow frequently accepted invalid records, highlighting that narrative plausibility may be an insufficient benchmark for the safe use of synthetic EHR data. Coogee addresses this by explicitly separating the two concerns: the generator handles probability, while the auditor acts as a proxy for clinical reasoning. This modularity ensures that the final output aligns not only with historical data distributions, but also with the logic of medical practice. Nonetheless, in future work it may be possible to merge the generative and auditing steps by leveraging recent advancements in the use of reinforcement learning for training generative models.\citepunc{guo2025deepseek}

While high-fidelity synthetic data is increasingly proposed as a surrogate for real-world evidence, we caution that it remains a proxy of historical practice. Models trained on such data may not reflect contemporary clinical guidelines or evolving best practices.\citepunc{kraljevic2024foresight}
Similarly, Coogee cohorts inherently reflect the biases and care patterns of the source training data, i.e., in this case, tertiary care from MIMIC-IV database. Consequently, these findings should be validated using diverse international datasets, including, for example, primary care records, to ensure generalizability.

This study establishes a foundation for clinically consistent synthetic data, yet several avenues remain for technical and translational advancement. Future work should explore: (1) \emph{Lifelong patient modeling}, adopting emerging architectures (e.g., state-space models) to generate multi-decade trajectories and overcome small fixed context window limitations;\citepunc{gu2021efficiently} (2) \emph{Retrieval-Augmented Auditing}, connecting the LLM auditor to external knowledge bases (e.g., PubMed, clinical guidelines) to verify complex decisions beyond its internal weights; (3) \emph{Human-in-the-loop refinement}, using clinician feedback to fine-tune the auditor's sensitivity to specific domain errors; and (4) \emph{Expansion to multimodal data}, incorporating unstructured clinical notes and imaging metadata for a more holistic patient representation.

In summary, Coogee bridges the gap between realistic generation and plausible clinical reasoning. It serves as a foundational tool for digital health, enabling sovereign sharing of high-quality, privacy-preserving patient data for clinical AI development and reproducible science.

\section*{Data sharing}
\noindent MIMIC-IV (v2.2) is available to credentialed researchers via PhysioNet (\url{https://physionet.org/content/mimiciv/2.2/}). The PrimeKG knowledge graph is available at \url{https://github.com/mims-harvard/PrimeKG}. All source code for the Coogee framework is provided at \url{https://github.com/jameszhou-gl/Coogee}.

\printcredits

\bibliographystyle{elsarticle-num}
\bibliography{cas-refs}

\clearpage

\renewcommand{\thefigure}{S\arabic{figure}}
\setcounter{figure}{0}
\renewcommand{\thetable}{S\arabic{table}}
\setcounter{table}{0}

\phantomsection
\addcontentsline{toc}{section}{Supplementary Appendix}
\section*{Supplementary Appendix}
\label{837154266846}
\setcounter{subsection}{0}
\renewcommand{\thesubsection}{S\arabic{subsection}}
\renewcommand{\theHsubsection}{Appendix.\arabic{subsection}}

\subsection{Data sources}
\label{796598494627}
We utilized the Medical Information Mart for Intensive Care (MIMIC-IV) database (v2.2), a comprehensive, de-identified EHR dataset sourced from the Beth Israel Deaconess Medical Center (Boston, MA, USA).\citepunc{johnson2023mimic} 
Access to the database was granted following completion of the required credentialing and ethics training via PhysioNet.
To capture the full spectrum of longitudinal patient care, we extracted data from five primary clinical domains. Static demographic variables such as sex, birth year, and ethnicity, were retrieved from the \emph{patients} table, while admission timing and metadata were obtained from the \emph{admissions} table. Clinical events were mapped to standard ontologies to ensure reproducibility: diagnoses were extracted from \emph{diagnoses\_icd} (International Classification of Diseases, Tenth Revision, Clinical Modification, ICD-10-CM), procedures from \emph{procedures\_icd} (ICD-10-PCS), and medication prescriptions from the \emph{prescriptions} table (mapped to Anatomical Therapeutic Chemical, ATC, codes using established conversion tables).\citepunc{renc2024zero} Laboratory results were obtained from \emph{labevents} and linked to their names using the \emph{d\_labitems} table.

\begin{table*}[hb]
\centering
\caption{\textbf{Patient characteristics in the MIMIC-IV dataset used in our experiments.} The data distribution is consistent across splits, ensuring comparability of training and evaluation cohorts.}
\label{816874121706}
\resizebox{\textwidth}{!}{%
\begin{tabular}{@{}llcccc@{}}
\toprule
 &  & \textbf{Train} & \textbf{Validation} & \textbf{Test} & \textbf{Total} \\ \midrule
\multirow{3}{*}{\textbf{Sample size, n}} & Patients & 146,377 (81\%) & 16,264 (9\%) & 18,071 (10\%) & 180,712 \\
& Hospital admissions &  348,900 & 39,432 & 42,743 & 431,075\\ 
 & Tokens & 121,424,363 & 13,665,709 & 14,467,629 & 149,557,701\\ \midrule
\textbf{Age in years, mean (std)} & - & 56$\cdot$20 (20$\cdot$19) & 56$\cdot$21 (20$\cdot$08) & 56$\cdot$36 (20$\cdot$27) & 56$\cdot$22 (20$\cdot$19) \\ \midrule
\multirow{3}{*}{\textbf{Gender, n (\%)}} & Female & 77,344 (52$\cdot$8\%) & 8,719 (53$\cdot$6\%) & 9,656 (53$\cdot$4\%) & 95,719 (53$\cdot$0\%) \\
 & Male & 46,202 (45$\cdot$4\%) & 7,545 (46$\cdot$4\%) & 8,415 (46$\cdot$6\%) & 84,993 (47$\cdot$0\%)\\ \midrule
\multirow{6}{*}{\textbf{Ethnicity, n (\%)}} & White & 97,664 (66$\cdot$7\%) & 10,768 (66$\cdot$2\%) & 12,010 (66$\cdot$5\%) & 120,442 (66$\cdot$6\%)\\
& Black & 18,735 (12$\cdot$8\%) & 2,084 (12$\cdot$8\%) & 2,382 (13$\cdot$2\%) & 23,201 (12$\cdot$8\%)\\
& Hispanic/South American & 8,105 (5$\cdot$5\%) & 934 (5$\cdot$7\%) & 1,010 (5$\cdot$6\%) & 10,049 (5$\cdot$6\%)\\
& Asian & 6,028 (4$\cdot$1\%) & 679 (4$\cdot$2\%) & 733 (4$\cdot$1\%) & 7,440 (4$\cdot$1\%)\\
& Other & 6,701 (4$\cdot$6\%) & 787 (4$\cdot$8\%) & 815 (4$\cdot$5\%) & 8,303 (4$\cdot$6\%)\\ 
 & Unknown & 9,144 (6$\cdot$2\%) & 1012 (6$\cdot$2\%) & 1,121 (6$\cdot$2\%) & 11,277 (6$\cdot$2\%)\\ \midrule
\multirow{3}{*}{\textbf{Marital status, n (\%)}} & Married & 63,280 (43$\cdot$2\%) & 7,058 (43$\cdot$4\%) & 7,733 (42$\cdot$8\%) & 78,071 (43$\cdot$2\%)\\
 & Single & 54,822 (37$\cdot$5\%) & 6,088 (37$\cdot$4\%) & 6,788 (37$\cdot$6\%) & 67,698 (37$\cdot$5\%)\\
 & Widowed & 12,858 (8$\cdot$8\%) & 1,419 (8$\cdot$7\%) & 1,635 (9$\cdot$0\%) & 15,912 (8$\cdot$8\%)\\
 & Divorced & 9,046 (6$\cdot$2\%) & 1007 (6$\cdot$2\%) & 1,121 (6$\cdot$2\%) & 11,174 (6$\cdot$2\%)\\
 & Unknown & 6,371 (4$\cdot$4\%) & 692 (4$\cdot$3\%) & 794 (4$\cdot$4\%) & 7,857 (4$\cdot$3\%)\\
\bottomrule
\end{tabular}%
}
\end{table*}

\begin{figure*}[!tbh]
\centering
\includegraphics[width=1.0\linewidth,keepaspectratio]{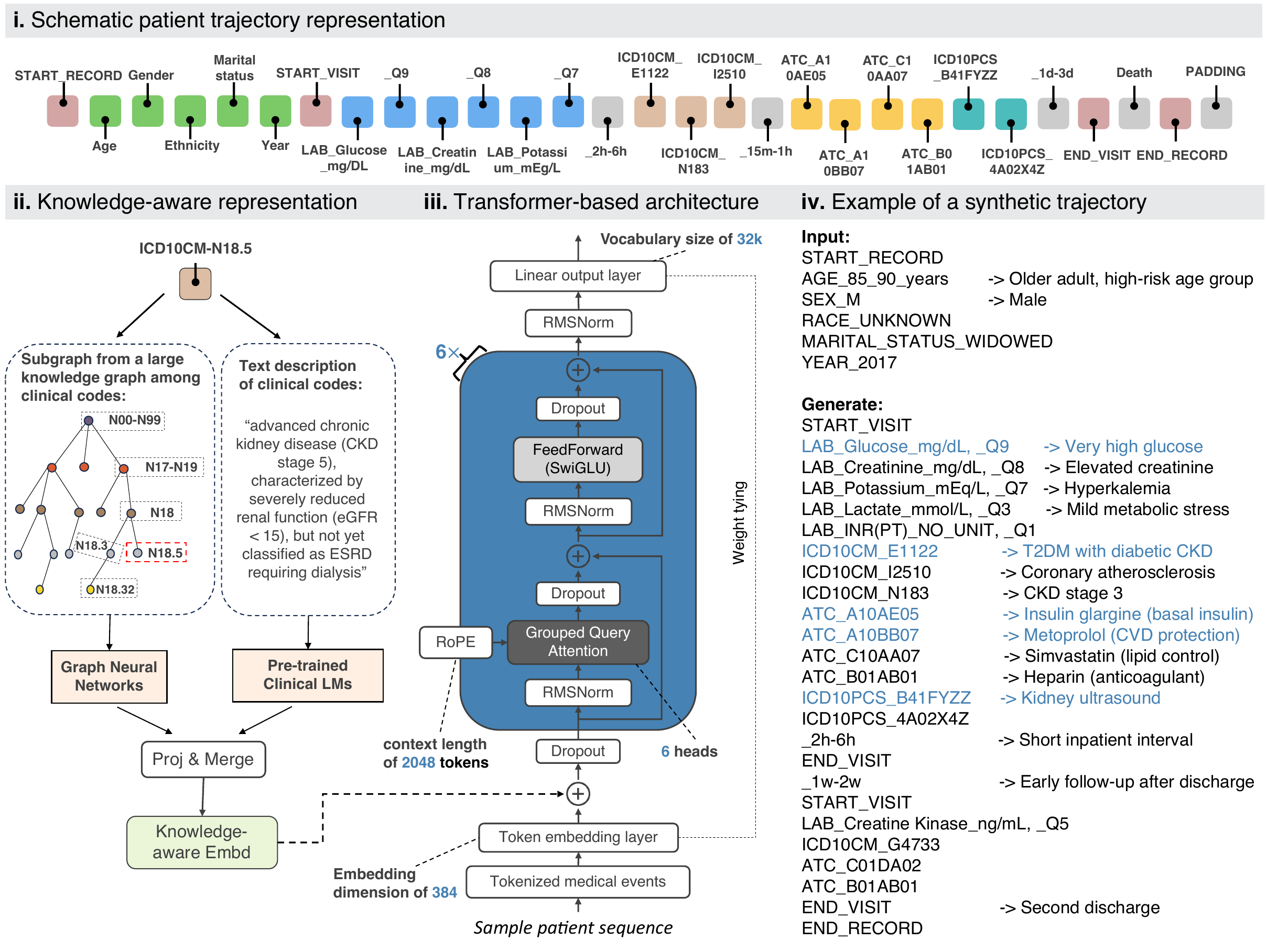}
\caption{\textbf{Generating high-fidelity, full-spectrum patient trajectories with Coogee.} 
\textbf{i}, Schematic patient trajectory representation integrating multimodal information: demographics (age, sex, race, marital status, calender year), laboratory tests, diagnoses (ICD-10-CM), procedures (ICD-10-PCS), medications (ATC codes), and structural tokens that encode time gaps and delimiters for records and visits). 
\textbf{ii}, Knowledge-aware representation of clinical events: each code is linked to its ontology hierarchy (e.g., ICD structure) and mapped to textual descriptions.
\textbf{iii}, Transformer-based generative architecture: tokenized medical events are embedded, enriched with knowledge-aware representations, and processed to model longitudinal and multimodal patient trajectories. 
\textbf{iv}, Example of a synthetic trajectory generated by Coogee: starting from demographics, the model produces diabetes-related codes including labs (glucose, creatinine), diagnoses (type 2 diabetes with CKD), medications (insulin, metformin), and associated procedures, with realistic temporal gaps between visits.}
\label{overview_fig}
\end{figure*}

\subsection{Knowledge-grounded generative Transformer}
\label{206223520903}

\begin{table*}[htb]
\caption{\textbf{Atomic tokenization and vocabulary composition.} The Coogee vocabulary maps 31,901 distinct clinical concepts to single tokens to prevent code fragmentation. The schema integrates static demographics with longitudinal events, utilizing quantile bins to capture continuous laboratory values and discrete interval tokens to preserve temporal resolution.}
\label{828348301138}
\centering
\resizebox{0.99\textwidth}{!}{%
\begin{tabular}{l c l}
\toprule
Category & \# Unique Tokens & Description / Notes \\
\midrule
\multicolumn{3}{l}{\textit{Patient Attributes (Static)}} \\
Age                & 17    & Discretized 5-year intervals (e.g., \texttt{AGE\_15\_20\_years}, ..., \texttt{AGE\_95\_100\_years}) \\
Sex                & 2     & Biological sex (\texttt{SEX\_M} and \texttt{SEX\_F}) \\
Race/Ethnicity          & 6     & \texttt{RACE\_}\{\texttt{ASIAN, BLACK, HISPANIC, OTHER, UNKNOWN, WHITE}\}\\
Marital Status     & 5     & \texttt{MARITAL\_STATUS\_}\{\texttt{DIVORCED, MARRIED, SINGLE, UNKNOWN, WIDOWED}\} \\
Calendar Year      & 16    & Year of admission start (from \texttt{YEAR\_2005} to \texttt{YEAR\_2020}) \\
\midrule
\multicolumn{3}{l}{\textit{Clinical Events (Longitudinal)}} \\
Diagnoses          & 18,837 & ICD-10-CM codes (with ICD-9 mapped to ICD-10) \\
Procedures         & 11,360 & ICD-10-PCS interventional procedure codes \\
Medications        & 1,167  & Drugs mapped to ATC codes \\
Laboratory Tests   & 461   & Distinct laboratory test identifiers (e.g., \texttt{LAB\_Glucose\_mg/dL}) \\
Lab Values      & 10    & Decile bins (\texttt{\_Q1}–\texttt{\_Q10}) for discretizing continuous measurements\\
Mortality              & 1     & Explicit \texttt{DEATH} token indicating in-hospital mortality \\
\midrule
\multicolumn{3}{l}{\textit{Sequence Control \& Temporal Logic}} \\
Time Intervals          & 14    & Discrete gaps between events/visits (e.g., \texttt{\_<=5m}, \texttt{\_1d-3d}, \texttt{\_>6mt}) \\
Structure  & 5     & Hierarchy markers (\texttt{START\_RECORD}, \texttt{START\_VISIT}, \texttt{END\_VISIT}, \texttt{END\_RECORD}, \texttt{PADDING}) \\
\midrule
\textbf{Total Vocabulary}     & \textbf{31,901} & Total unique tokens modeled by Coogee \\
\bottomrule
\end{tabular}
}
\end{table*}

Electronic health records are characterized by high-dimensional, sparse vocabularies where many clinical codes appear infrequently (Figure~\ref{fidelity_ba}(a)). Purely data-driven models often fail to learn robust representations for these ``long-tail'' concepts, treating them as arbitrary symbols. To resolve this, we construct knowledge-grounded embeddings that enrich each token with two complementary layers of biomedical context: (i) structural knowledge derived from biological networks, and (ii) semantic knowledge derived from clinical definitions. 
 
We leverage PrimeKG, a precision-medicine knowledge graph containing over 129,000 nodes and 4 million relations aggregated from 20 curated resources across diseases, phenotypes, drugs, and biological pathways.\citepunc{chandak2023building} Using the MedToK resources, we map each clinical code (ICD-10-CM, ICD-10-PCS, and ATC) to its corresponding anchor nodes in the graph.\citepunc{su2025multimodal} To capture biological context, we encode the local neighborhood of each anchor node using a Relational Graph Convolutional Network (RGCN).\citepunc{schlichtkrull2018modeling} Instead of learning from statistical co-occurrence alone, the RGCN updates node representations by aggregating information from biologically related neighbors (pathways involved in a disease). We train this network using an edge reconstruction objective, forcing the model to assign similar representations to clinically related concepts even if they rarely appear together in patient records. The final structural embedding is obtained by pooling these learned representations.

Let $\mathcal{G}=(\mathcal{N}, \mathcal{R}, \mathcal{E})$ denote the knowledge graph, where $\mathcal{N}$ is the set of nodes, $\mathcal{R}$ is the set of relations, and $\mathcal{E}$ is the set of edges. For a clinical code $c$, we identify a set of anchor nodes $A_c \subseteq \mathcal{N}$ via ontological mapping. We induce a subgraph $\mathcal{G}_c$ containing $A_c$ and their $k$-hop neighbors. We employ a Relational Graph Convolutional Network (RGCN) to learn node embeddings.\citepunc{schlichtkrull2018modeling}
Let $h_v^{(l)} \in \mathbb{R}^{d^{(l)}}$ denote the hidden state of node $v$ at layer $l$. The update rule aggregates messages from neighbors $\mathcal{N}_r(v)$ under relation $r \subset \mathcal{R}$ as follows:
\begin{equation} 
\label{913298302117}
h_v^{(l+1)} = \sigma \left( \sum_{r \in \mathcal{R}} \sum_{u \in \mathcal{N}_r(v)} \frac{1}{c_{v,r}} W_r^{(l)} h_u^{(l)} + W_0^{(l)} h_v^{(l)} \right) 
\end{equation}

where $\mathcal{N}_r(v)$ denotes the set of immediate neighbors of $v$ under relation $r$, $c_{v,r}$ is a normalization constant, $W_r^{(l)}$ and $W_0^{(l)}$ are learnable weight matrices, and $\sigma
(\cdot)$ is an element-wise activation function. The RGCN is optimized using a \emph{edge reconstruction loss} to preserve relational topology.
The raw structural embedding for code $c$, denoted $\hat{z}_c^{\text{struct}}$, is derived by mean-pooling the representations of its anchor nodes after $L$ layers:
\begin{equation}
    \label{29638107}
    \hat{z}_c^{\text{struct}} = \frac{1}{|A_c|}\sum_{v \in A_c} h_v^{(L)}
\end{equation}

In parallel, we capture the explicit clinical definition of each code by encoding its textual description (e.g., ``Type 2 diabetes mellitus without complications'' for the ICD-10-CM code E11.9) using ClinicalBERT.\citepunc{alsentzer2019publicly} This generates a semantic vector that captures linguistic nuances and hierarchical similarities inherent in medical nomenclature.
The raw semantic embedding $\hat{z}_c^{\text{sem}}$ is obtained by encoding the textual description $T(c)$ via ClinicalBERT and extracting the [CLS] token representation.

The final embedding for each clinical token is generated by projecting both the biological and linguistic signals into the Transformer's latent dimension and summing them. For the small subset of the vocabulary lacking graph mappings, such as, demographics, lab tests, time intervals, and structural markers, the structural component is zeroed, and the representation relies solely on the projected semantic embedding. This ensures valid, high-fidelity inputs across the entire heterogeneous vocabulary.
To construct the final unified embedding, we first project both components independently into the Transformer's hidden dimension $d_{\text{model}}$ (Table~\ref{131977762643}):
\begin{align}
    h_c^{\text{struct}} &= W_{\text{struct}} \hat{z}_c^{\text{struct}} + b_{\text{struct}} \\
    h_c^{\text{sem}} &= W_{\text{sem}} \hat{z}^{\text{sem}} + b_{\text{sem}}
\end{align}
where $W_{\text{struct}} \in \mathbb{R}^{d_{\text{model}} \times d^{(L)}}$ and $W_{\text{sem}} \in \mathbb{R}^{d_{\text{model}} \times d_{\text{BERT}}}$ are learnable projection matrices.
The final embedding $z_c \in \mathbb{R}^{d_{\text{model}}}$ is the layer-normalized sum of these projected components:
\begin{equation}
    z_c = \text{LayerNorm}\left( h_c^{\text{struct}} + h_c^{\text{sem}} \right)
\end{equation}
For tokens lacking structural mappings ($A_c=\emptyset$), we set $\hat{z}_c^{\text{struct}}=\bm{0}$, so that the representation relies solely on the semantic component.

We instantiated Coogee using a decoder-only Transformer architecture, optimized for autoregressive sequence modeling.\citepunc{vaswani2017attention,radford2019language}
As shown in Figure~\ref{overview_fig}, the model processes patient trajectories as a sequence of tokens, predicting the next clinical event based on the entire preceding history. To handle the complexity of longitudinal patient care, we incorporated architectural advances from recent large language models to improve stability and temporal reasoning:\citepunc{agarwal2025gpt,yang2025qwen3} 
\begin{enumerate*}[label=(\roman*)]
    \item \textbf{Rotary Position Embeddings (RoPE)}.\citepunc{su2024roformer} We replaced standard absolute position encodings with RoPE. This allows the model to better generalize to varying sequence lengths and capture relative temporal distances between clinical events (e.g., the gap between a diagnosis and a procedure).
    \item \textbf{Efficient Attention and Activations.} We utilized Grouped-Query Attention (GQA) to reduce memory overhead during inference and SwiGLU activations in the feed-forward networks to enhance non-linear modeling capabilities.\citepunc{ainslie2023gqa,shazeer2020glu}
    \item \textbf{Pre-Normalization.} To stabilize training at deeper layers, we applied RMSNorm before each attention and feed-forward block.\citepunc{zhang2019root}
    \item \textbf{Embedding Factorization.} To reduce the parameter footprint of the large clinical vocabulary, we decomposed the embedding matrix. Instead of projecting the 31,901 tokens directly into the hidden dimension ($d=384$), we map them to a lower-dimensional space ($d=100$) first, and then project to the full hidden dimension. This factorization reduces approximately 9 million trainable parameters without compromising representation capacity. 
\end{enumerate*}
The final architecture consists of 6 decoder layers with a hidden dimension of 384, 6 attention heads, and a context window of 2,048 tokens. The total model size is approximately 50M parameters. (See Supplementary Table~\ref{131977762643} for a detailed parameter breakdown.)

\begin{figure*}[!htb]
\centering
\includegraphics[width=\linewidth,keepaspectratio]{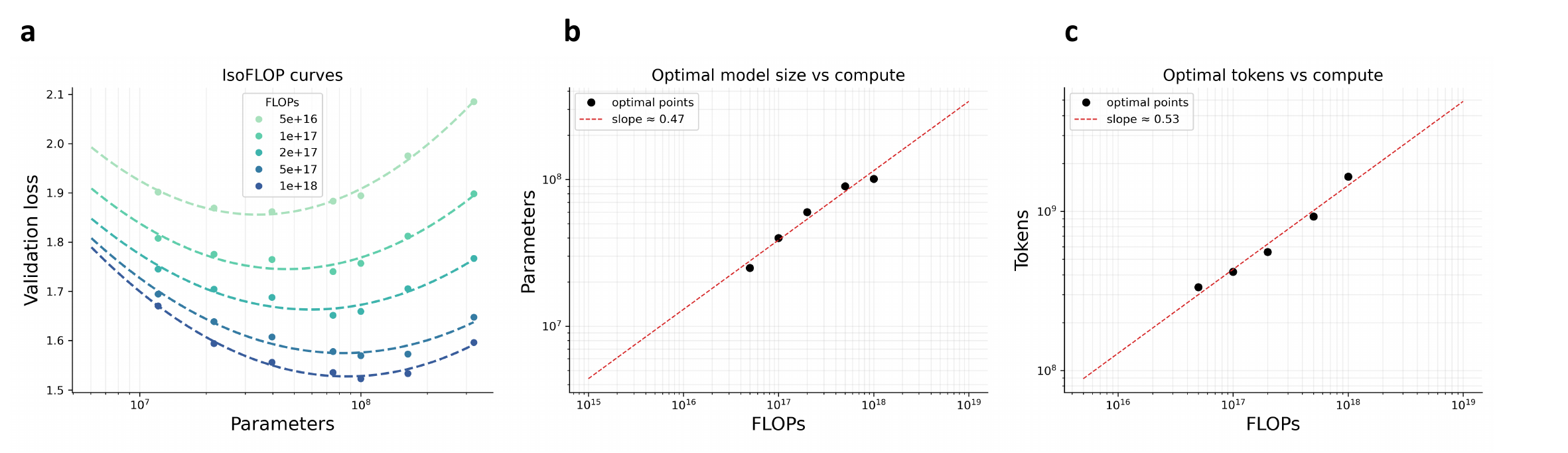}
\caption{\textbf{Scaling law investigation of model-data trade-offs in Coogee.} (a) IsoFLOP curves showing validation loss as a function of model size (parameters) for different computational budgets. Each curve corresponds to a constant training FLOPs constraint, with dashed lines denoting quadratic fits in log–parameter space. (b) Optimal model size (minimizing validation loss per compute budget) follows a power-law relationship with total training FLOPs, consistent with Chinchilla-style scaling trends.\citepunc{hoffmann2022training} (c) Corresponding optimal token counts also obey a power-law with compute. Together, these analyses quantify the balance between model capacity and dataset size for efficient training. Based on the identified isoFLOP frontier and available computational resources, we selected the 50M-parameter (13M trainable) configuration for our main experiments.
}
\label{971551448872}
\end{figure*}

\subsection{Scaling laws and compute-optimal model selection}
\label{216137144624}
\noindent We conducted a scaling law investigation to identify the compute-optimal allocation of parameters and data for the Coogee framework and to ensure efficient training. Following recent methodologies for large-scale modeling\citepunc{hoffmann2022training,zhang2025exploring}, we analyzed the trade-offs between model capacity and dataset size across five computational budgets, ranging from $5\times10^{16}$ to $1\times10^{18}$ FLOPs.

As shown in Figure~\ref{971551448872}, our IsoFLOP curves exhibit a clear frontier where validation loss is minimized for each specific budget. By fitting these results to power-law relationships, we observed that both optimal model size (slope $\approx$0$\cdot$47) and optimal token counts (slope $\approx$0$\cdot$53) scale predictably with increased compute. This analysis determines the selection of the 50M-parameter configuration (13M trainable) based on our available resources.

\subsection{Training, inference, and implementation details}
\label{617393837400}

\begin{table*}[thb]
\centering
\caption{\textbf{Architecture and parameter breakdown of the Coogee-50M model.} The table details the hyperparameters, embedding dimensions, and per-component parameter counts. Of the total 50M parameters, 13.2M are trainable and 36.7M are frozen. Knowledge-aware embeddings, including structural (hierarchy and semantic) embeddings, are frozen to preserve ontology-derived structure, while trainable parameters lie primarily in the token embedding layer and decoder blocks.}
\label{131977762643}
\begin{tabular}{lcrrr}
\toprule
\textbf{Hyperparameter} & \textbf{Value} & \textbf{Params} & \textbf{Trainable} & \textbf{Frozen} \\
\midrule
Vocab size        & 31,901 & – & – & – \\
Context length    & 2,048 & – & – & – \\
Embedding dim & 384 & – & – & – \\
Hidden dim        & 1,024 & – & – & – \\
Decoder layers    & 6 & – & – & – \\
Attention heads   & 6 & – & – & – \\
Key/Value heads   & 2 & – & – & – \\
Batch size        & 32 & – & – & – \\
\midrule
\multicolumn{5}{l}{\textbf{Parameter breakdown}} \\ \midrule
Token Embeddings        & [31,901 $\times$ 100] & 3,190,100 & 3,190,100 & 0 \\ 
Token Embeddings Projection        & [384 $\times$ 100] & 38,400 & 38,400 & 0 \\ 
Hierarchy Embeddings    & [31,901 $\times$ 384] & 12,249,984 & 0   & 12,249,984\\
Hierarchy Projection & [384$\times$384] & 147,456 & 147,456 & 0\\
RMSNorm & [384] & 384 & 384 & 0\\
alpha\_h & - & 1 & 1 & 0\\ 
Semantic Embeddings     & [31,901 $\times$ 768] & 24,499,968 & 0    & 24,499,968 \\
Semantic Projection & [768$\times$384] & 294,912 & 294,912 & 0\\
RMSNorm & [384] & 384 & 384 & 0\\
alpha\_s & - & 1 & 1 & 0\\ 
DecoderBlock (6 $\times$) & $n_{\text{embd}}=384, n_{\text{hid}}=1024, n_{\text{heads}}=6$ & 9,441,792 & 9,441,792 & 0 \\
Final norm              & [384] & 384 & 384 & 0 \\
Output Projection & - & 38,400 & 38,400 & 0\\
\midrule
\textbf{Total}          & – & \textbf{49,902,166} & \textbf{13,152,214} & \textbf{36,749,952} \\
\bottomrule
\end{tabular}
\end{table*}

\noindent The model was trained using a causal language modeling objective. At each time step $t$, the model minimizes the negative log-likelihood of the next token $x_{t+1}$ given the history $x_{0:t}$. This forces the model to learn both the statistical prevalence of medical codes and the sequential logic of clinical care. We utilized the AdamW optimizer with betas $(0.9, 0.95)$ and an initial learning rate of $3\times10^{-4}$. We applied a schedule with $1\%$ linear warmup steps followed by cosine decay down to 10\% of the peak learning rate.\citepunc{hoffmann2022training} To prevent overfitting on the training cohort, we applied a weight decay $0.1$, dropout of $0.1$, and early stopping based on validation loss (patience = 10).
Training was conducted on a single NVIDIA H200 GPU for 200 epochs.

\noindent To generate synthetic patient records, we initialize the model with the \texttt{START_RECORD} token and the static demographic attributes from the test cohorts to establish a baseline patient profile. Subsequently, the model enters a recursive longitudinal cycle, generating discrete clinical encounters delimited by \texttt{START_VISIT} and \texttt{END_VISIT} markers. 
Within this loop, the model autonomously determines the temporal cadence of care by predicting specific time-interval tokens (e.g., \texttt{_3m-6m}) between events. 
To ensure the structural integrity of laboratory data, we employ constrained decoding: whenever the model generates a laboratory test token (e.g., \texttt{LAB_HbA1c}), the sampling distribution for the subsequent step is strictly restricted to valid quantile tokens (e.g., \texttt{_Q}). Conversely, quantile tokens are masked out in all other contexts to prevent the generation of invalid values.
To ensure the synthetic cohort reflects the diversity of the real population while filtering out implausible low-probability events, we employ nucleus sampling ($\text{top-p}=0.98$, $\text{temperature}=1.0$) throughout the decoding process.\citepunc{holtzman2019curious}
For all discretized tokens representing continuous data, specifically time intervals and laboratory quantiles, we apply a post-processing step that samples a precise numerical value uniformly from the token's defined range. For instance, we may convert \texttt{_3m-6m} token into a specific duration such as 142 days.
This generation continues until the model produces an \texttt{END_RECORD} or \texttt{DEATH} token.

We implemented Coogee using PyTorch 2.5.0+cu121, along with scikit-learn 1.5.2, NumPy 2.1.2, and transformers 4.46.2.  
The model was trained on a single NVIDIA H200 GPU. 
Table~\ref{131977762643} details the specific architectural hyperparameters. Note that the total parameter count is approximately 50M, as informed by the scaling law analysis in Appendix~\ref{216137144624}.
Only 13M parameters are trainable, because the knowledge-grounded embeddings (Structural and Semantic) are pre-computed and frozen during training to preserve the rigorous ontological structure derived from PrimeKG and ClinicalBERT. The model learns to project and gate these frozen representations rather than modifying them.

\subsection{Auditing implementation details}
\label{941068451355}
\noindent To ensure reproducible evaluation, we utilized the following structured system prompt for all LLM evaluators. The prompt enforces a persona (Chief Medical Officer), defines strict evaluation dimensions, and mandates a structured JSON output to enable parsing (Listing~\ref{146058716182}).

\begin{lstlisting}[
language=prompt, 
caption={Full system prompt used for automated fidelity evaluation.}, 
label={146058716182}
]
You are an expert Chief Medical Officer and Data Quality Auditor. You are reviewing a patient record for clinical and logical validity.

### DATA FORMAT
You will be provided with a CSV content representing a patient's timeline.
- **time**: ISO 8601 timestamp. Events are chronological.
- **code**: The raw clinical identifier (e.g., ICD10CM, ICD10PCS, ATC, LAB test, Demographic tags like `SEX_F').
- **numerical_value**: The result for lab tests. NOTE: This field is empty/NaN for non-lab rows. Ignore empty values.
- **code_label**: Human-readable description of the code.

### EVALUATION CRITERIA
Analyze the record across three dimensions:
1. **Clinical Plausibility**: Do diagnoses, procedures, medications, and laboratory values make clinical sense together?
2. **Logical Consistency**: Check demographics (Age, Sex). Does a male patient have a pregnancy code? 
3. **Temporal Coherence**: Are events ordered and spaced in a clinically realistic timeframe?

### SCORING RUBRIC
1-2: Clearly artificial; repetitive patterns, impossible values, or severe logic breaks.
3-4: Largely synthetic with frequent errors or "hallucinations".
5-6: Plausible but has subtle inconsistencies (e.g., odd medication choices).
7-8: Mostly realistic; minor temporal oddities but clinically sound.
9-10: Indistinguishable from real-world EHR trajectories.

### PATIENT RECORD from the csv
[CSV Content Inserted Here]

### OUTPUT FORMAT
You must respond in the following JSON format:
{
  "realism_score": <INTEGER 1-10>,
  "reasoning": "A brief summary of your analysis, highlighting realistic vs. synthetic elements."
}
\end{lstlisting}

\subsection{Sensitivity analysis: impact of context window on fidelity}
\label{635022531020}
\noindent The main fidelity evaluation (Figure~\ref{fidelity_ba}) compares the synthetic cohort against the full, unfiltered real data. As noted in the main text, the fixed context window of the Transformer (2,048 tokens) inherently limits the generation of extremely long patient trajectories, resulting in a divergence in the upper tail of the code-count distribution.
To verify that this divergence is an architectural artifact rather than a failure of the generative logic, we performed a sensitivity analysis. We constructed a control set of real patient records restricted to those with a total token length $\leq$2,048 (representing 90$\cdot$8\% of the test population). 
We then re-evaluated the structural fidelity metrics against the synthetic cohort. 
As shown in Table~\ref{821578357459}, when controlling for context length, the statistical alignment becomes near-perfect. The difference in the mean number of codes per visit narrows to less than 2 codes, and the sequential probability distributions remain robust ($R^2>$0$\cdot$99). This confirms that Coogee correctly learns the underlying clinical distribution, and the deviations observed in the main text are attributable solely to the context truncation of extreme outliers.\citepunc{beltagy2020longformer}

\begin{table*}[h] 
\centering 
\caption{\textbf{Comparison of structural metrics controlled for context window limits.} When the real baseline is restricted to trajectories that fit within the model's context window ($\leq$ 2,048 tokens), the synthetic data mirrors the real distributions with high precision.} 
\label{821578357459} 
\begin{tabular}{lcc} \toprule 
\textbf{Metric} & \textbf{Real (Truncated)} & \textbf{Synthetic} \\ \midrule 
\multicolumn{3}{l}
{\textit{Trajectory Structure}} \\ Visits per Patient  & Median: 1$\cdot$0; IQI: 1$\cdot$0-2$\cdot$0 & Median:1$\cdot$0; IQI: 1$\cdot$0-3$\cdot$0 \\ Codes per Visit & Median:131$\cdot$0; IQI: 46$\cdot$0-316$\cdot$0  & Median:124$\cdot$0; IQI: 45$\cdot$0-289$\cdot$0  \\ 
\midrule 
\multicolumn{3}{l}{\textit{Probabilistic Alignment ($R^2$)}} \\ 
 Unigram Probability & \multicolumn{2}{c}{0$\cdot$997} \\ 
 Same-Visit Co-occurrence & \multicolumn{2}{c}{0$\cdot$988} \\ 
 Sequential-Visit Dependencies & \multicolumn{2}{c}{0$\cdot$992} \\ \bottomrule 
 \end{tabular} 
 \end{table*}

\subsection{Correlations of code probabilities between real and synthetic data}
\label{059936646241}

\begin{figure*}[!htb]
\centering
\includegraphics[width=0.96\linewidth,keepaspectratio]{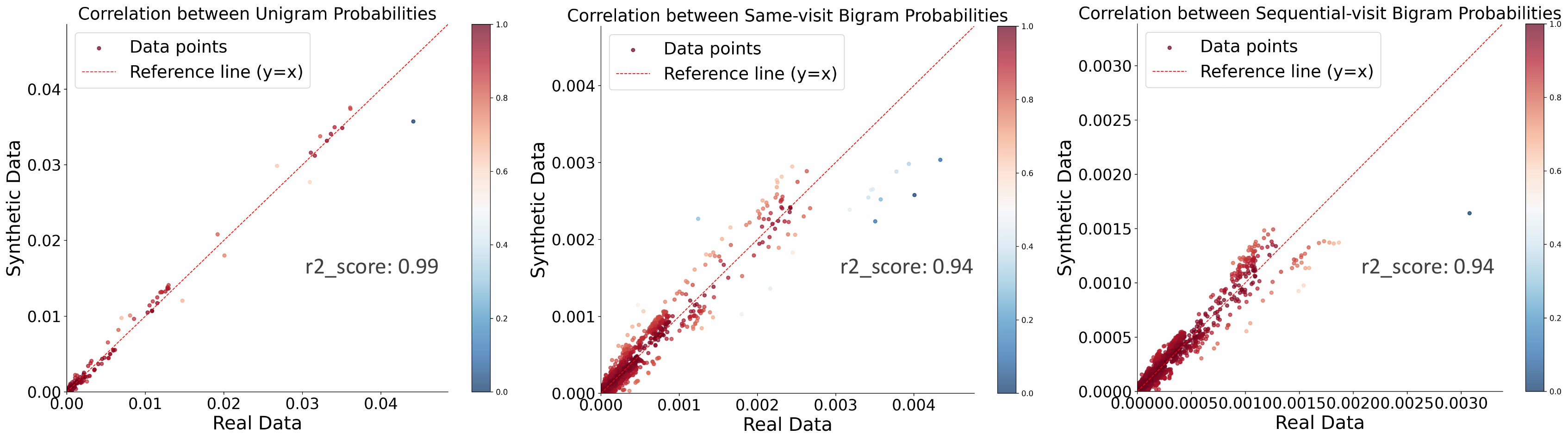}
\caption{Preservation of population-level statistics: correlations of unigram (marginal), same-visit bigram (co-occurrence), and sequential-visit bigram (temporal) probabilities.
}
\label{correlation}
\end{figure*}

\begin{figure*}[!th]
\centering
\includegraphics[width=0.93\linewidth,keepaspectratio]{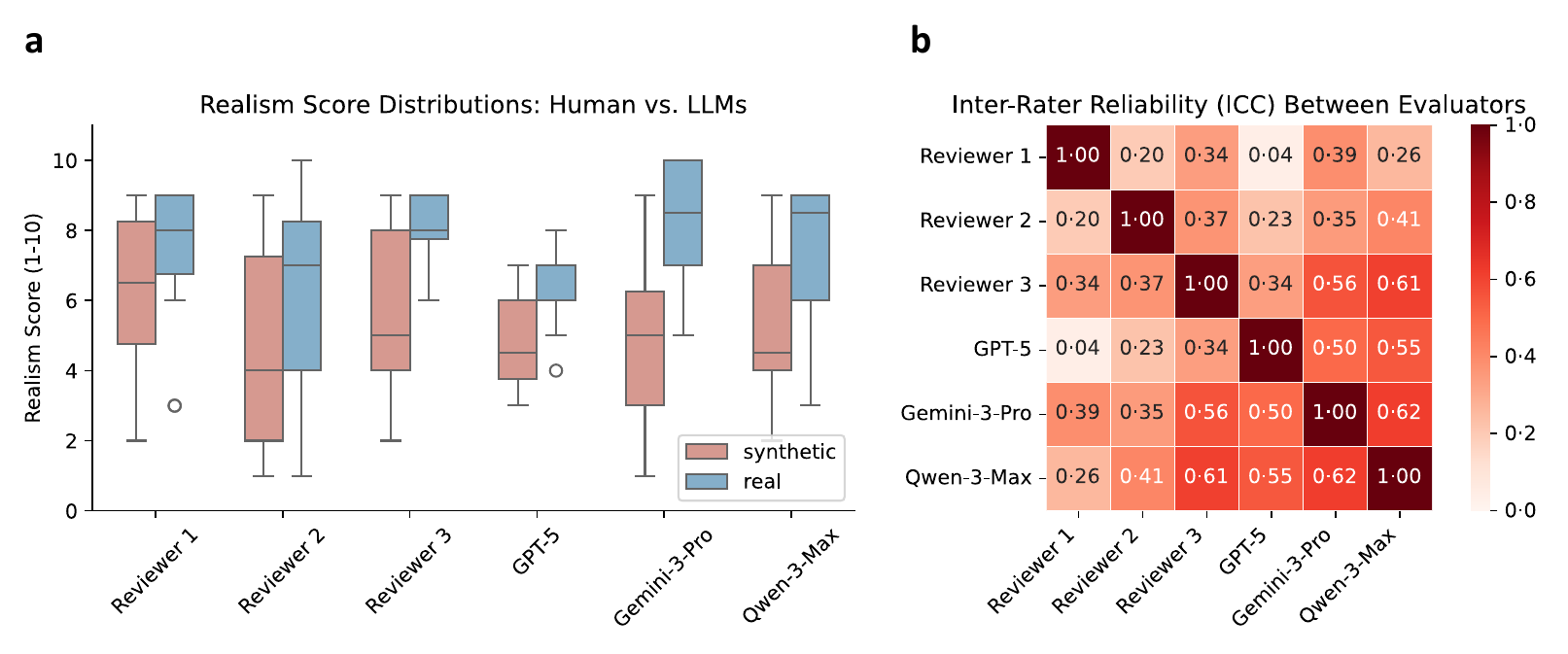}
\caption{\textbf{Clinician and LLM-based realism scores in Round 1.}
(a) Distribution of realism scores. Narrative-focused clinicians (Reviewers 1 and 2) show significant distributional overlap between Real (blue) and Synthetic (red) records. In contrast, the rule-checking clinician (Reviewer 3) and LLMs show a clear separation, identifying logical inconsistencies. (b) Inter-rater reliability measured through Intraclass Correlation Coefficients (ICCs). The strong correlation between Reviewer 3 and LLMs (ICCs of 0$\cdot$34, 0$\cdot$56 and 0$\cdot$61) suggests that automated auditors may focus on the validity of clinical rules within the records.
}
\label{realism-scores-round-1}
\end{figure*}

To quantify the validity of generated clinical patterns, we compared the association between event probabilities in real and synthetic data at three levels of granularity (Figure\ref{correlation}):
\begin{enumerate*}[label=(\roman*)]
\item \emph{Prevalence (Unigrams).} The marginal probabilities of individual codes showed near-perfect alignment (coefficient of determination $R^{2}=0$$\cdot$$99$), indicating that phenotype prevalence and medication prescription rates are faithfully reproduced. An $R^{2}$ close to 1 indicates that the variance in synthetic event probabilities is explained almost entirely by their real counterparts.
\item \emph{Co-occurrence (Same-visit Bigrams).} The model preserved dependencies within visits ($R^{2}=0$$\cdot$$94$), correctly associating diagnoses with their relevant medications and procedures. 
\item \emph{Care pathways (Sequential Bigrams).} Longitudinal dependencies across visits achieved similar high fidelity ($R^{2}=0$$\cdot$$94$), suggesting that the model captures the temporal sequence of care (e.g., a diagnosis in Visit 1 leading to associated procedures in Visit 2).
\end{enumerate*}

\subsection{Efficient factorization}
\label{045304954792}
We recognize that avoiding code-splitting increases vocabulary size and consequently the parameter burden of the token embedding layer. In our final Coogee model, the token embedding layer needs 12M parameters and constitutes 55\% of overall trainable parameters (Table~\ref{131977762643}), scaling linearly with vocabulary size. 
To address this trade-off, we adopt a matrix factorization strategy inspired by ALBERT, decomposing the original embedding matrix $E_{\text{tok}} \in \mathbb{R}^{V\times D}$ into two smaller matrices $E_{1} \in \mathbb{R}^{V\times E}$ and $E_{2} \in \mathbb{R}^{E\times D}$, where $E\ll D$.\citepunc{lan2019albert}
Setting $E=100$ reduced the embedding parameters from 31,901 $\times$ 384 to 31,901 $\times$ 100 + 100 $\times$ 384, corresponding to a \emph{3.6$\times$ reduction} in the embedding layer size. Despite this compression, model fidelity remained nearly unchanged. The reconstruction performance showed minimal variation: ($R^2_{\text{uni}}$: 0.99 $\rightarrow$ 0.99; $R^2_{\text{same}}$:0.93 $\rightarrow$ 0.92; $R^2_{\text{seq}}$:0.92 $\rightarrow$ 0.92).
Predictive utilities were also comparable across downstream tasks: AUROC for phenotype (0.914 $\rightarrow$ 0.923), mortality (0.704 $\rightarrow$ 0.665), length-of-stay (0.734 $\rightarrow$ 0.724), and readmission (0.571 $\rightarrow$ 0.585). These findings indicate that factorizing the token embedding into lower-rank components can substantially reduce model parameters with negligible loss of representational power, supporting the extension of Coogee's tokenization method to larger vocabularies or longer contexts under constrained compute budgets.

\begin{figure*}[H]
\centering
\includegraphics[width=0.8\linewidth,keepaspectratio]{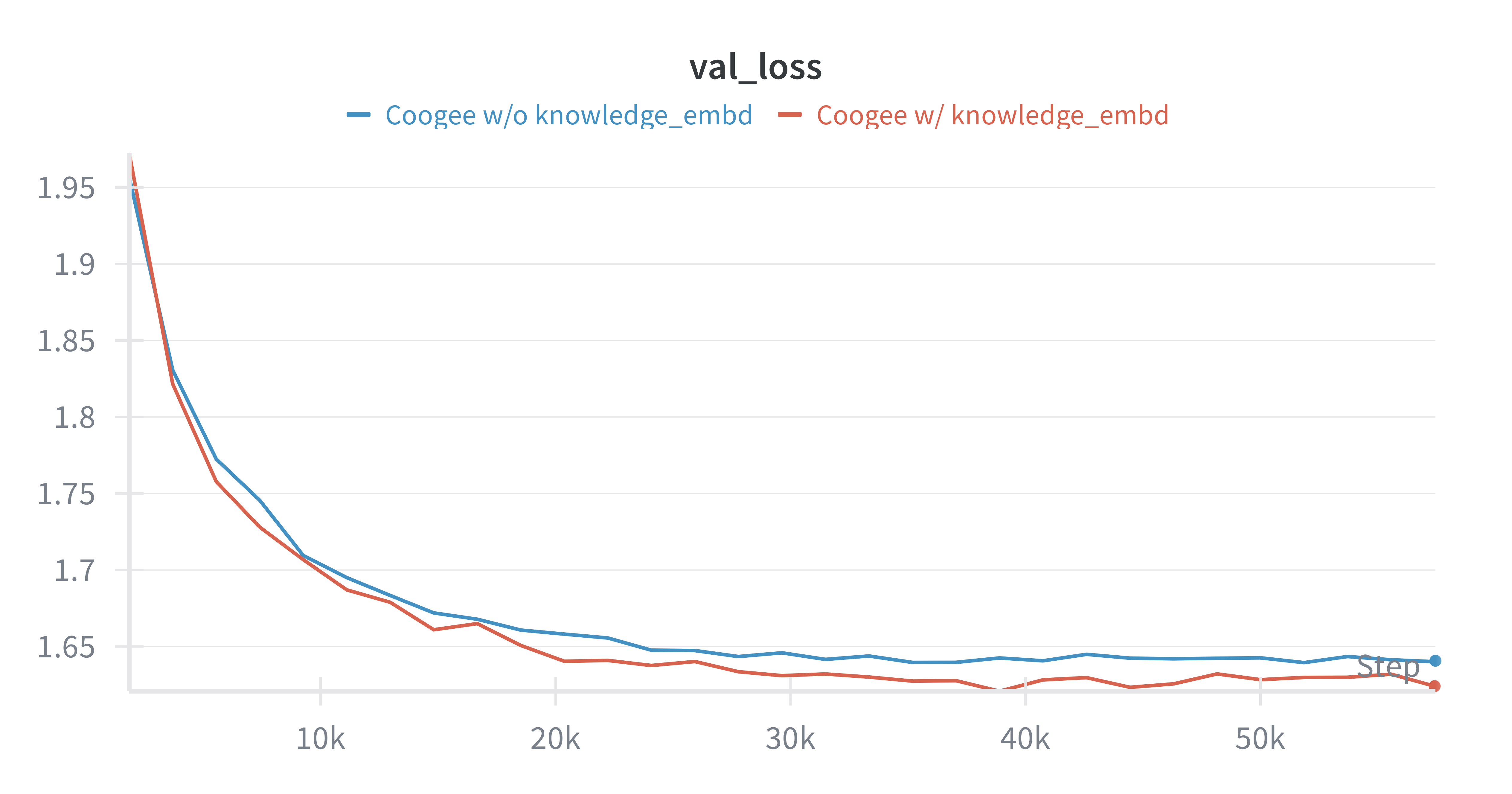}
\caption{\textbf{Validation loss curves of Coogee with and without knowledge-informed embeddings.} The plot compares the validation loss trajectories during training for two model variants: Coogee with \texttt{knowledge\_embd} (red) and Coogee without \texttt{knowledge\_embd} (blue). Both models share identical hyperparameters and training configurations, differing only in the inclusion of auxiliary biomedical knowledge embeddings (hierarchical and semantic). The model with knowledge integration exhibits consistently lower validation loss across training steps.
}
\label{val-loss}
\end{figure*}

\subsection{Impact of knowledge-grounded AI}

An important contribution of this work is the incorporation of knowledge-aware embeddings into the generative framework. Conventional approaches that rely solely on token embeddings treat each medical code as an independent symbol, without leveraging the ontological and semantic structure that underlies EHR vocabularies. This can limit their ability to capture relationships between clinically related codes, especially for less frequent events. By integrating knowledge-aware embeddings, Coogee is able to propagate semantic information across related diagnoses, procedures, and medications, thereby converging to a lower validation loss (Figure~\ref{val-loss}) and improving the fidelity of generated patient trajectories. To quantify this effect, we conducted an ablation study comparing Coogee with and without knowledge-aware embeddings. Without knowledge integration, alignment with real data distributions was reduced, with $R^{2}$ scores of 0.981 for unigrams, 0.899 for same-visit bigrams, and 0.909 for sequential bigrams. Incorporating knowledge-aware embeddings improved these to 0.990, 0.930 and 0.930, corresponding to absolute gains of 0.9–3.1 percentage points, particularly strengthening co-occurrence structures. Gains were also evident in downstream predictive tasks under the TSTR protocol. Mortality prediction improved from AUROC 0.689 (95\% CI: 0.658–0.721) to 0.711 (95\% CI: 0.675–0.742), a +2.2 \% point increase. Length-of-stay rose from 0.725 (95\% CI: 0.710–0.741) to 0.734 (95\% CI: 0.718–0.750), while readmission increased from 0.556 (95\% CI: 0.523–0.591) to 0.581 (95\% CI: 0.547–0.616), reflecting consistent 2–3 \% point gains.

A central distinction between Coogee and recent generative Transformer models for EHR data such as ETHOS and Curiosity is that the latter were not designed with synthetic data generation as their primary goal.\citepunc{renc2024zero,waxler2025generative} Both adopt code-splitting tokenization strategies, decomposing a single medical concept into multiple sub-tokens to reduce vocabulary size and ease language-model training. While effective for predictive tasks, this design introduces challenges for generation, as sub-tokens can recombine into invalid or clinically uninterpretable codes. Instead, Coogee preserves a one-to-one mapping between tokens and ontology-defined events. This choice has two important implications. First, it guarantees validity: every generated token corresponds directly to a recognized diagnosis, procedure, medication or lab concept, eliminating recombination errors that accumulated at scale. For example, Curiosity reports an invalid token ratio of 0.02\%; while seemingly small, this translates into tens of thousands of invalid codes in our patient cohort size. By contrast, Coogee maintains validity across the full space of 32k clinical concepts, an order of magnitude broader than prior works in synthetic EHR data generation. Second, the concept-level vocabulary supports interpretable representations and enables seamless integration with knowledge-aware embeddings, ensuring that generated sequences remain clinically coherent and ontology-aligned.

\subsection{Related work}
\textbf{Synthetic EHR data generation.}
The synthesis of EHRs has been explored through several methodological paradigms.\citepunc{chen2024generating} Early rule-based approaches rely on predefined clinical guidelines to simulate patient trajectories, ensuring interpretability but failing to capture statistical complexity of real-world data.\citepunc{buczak2010data,franklin2014plasmode,mclachlan2018aten}
GAN-based models generate synthetic records by matching distributions between real and synthetic samples,\citepunc{zhang2020ensuring,yale2020generation,wang2024igamt,li2023generating,baowaly2019synthesizing,yang2019grouped,lee2020generating,yan2021generating,torfi2020corgan,kuo2022health} while VAE-based methods learn continuous latent representations for stochastic sampling.\citepunc{biswal2021eva,sun2024collaborative}
However, both GAN- and VAE-based methods struggle with the sparse and long-tailed nature of medical codes and often fail to capture fine-grained dependencies.
More recently, Transformer and Diffusion-based generators have been introduce to model patient records.\citepunc{kuo2023synthetic,tian2024reliable}
HALO and HiSGT extend synthetic EHR generation to vocabularies of thousands of medical codes.\citepunc{theodorou2023synthesize,Zhou2025-mt} Yet, these approaches remain restricted to specific tabular domains, such as diagnoses only, thus do not model the full spectrum of the patient sequences.

\textbf{Generative Transformer.}
The Transformer architecture has revolutionized sequence modeling in natural language processing,\citepunc{Vaswani2017AttentionIA,radford2019language,brown2020languagemodelsfewshotlearners} computer vision,\citepunc{dosovitskiy2021imageworth16x16words,zhou2024hcvp} and multimodal learning.\citepunc{yang2023dawn,han2024well,zhou2026small} In healthcare, Transformers have been adapted for generative and forecasting tasks over longitudinal EHRs. ETHOS and Curiosity leverage generative Transformers to simulate future health trajectories and achieve strong performance on predictive tasks.\citepunc{renc2024zero,waxler2025generative} However, their tokenization strategies are optimized for trajectory forecasting rather than sequence integrity, often fragmenting clinical codes into multiple tokens. While this design supports prediction tasks, it introduces invalid or clinically implausible events in synthetic generation, making them unsuitable for high-fidelity data synthesis.

\onecolumn
\subsection{Example of synthetic patient trajectory}
\label{942767755247}

\begin{longtable}{@{}p{0.1\textwidth}p{0.36\textwidth}p{0.16\textwidth}p{0.3\textwidth}@{}}
\caption{\textbf{A synthetic patient trajectory achieving a mean realism score of 7.7/10 in combined clinician and automated evaluation (\S\ref{944464728637}).} 
This synthetic record represents an elderly female undergoing Coronary Artery Bypass Grafting (CABG). 
Evaluators selected this trajectory for its high degree of pathophysiological coherence. 
Specifically, the human reviewer noted that the severe drop in hemoglobin and hematocrit was physiologically consistent with post-cardiac surgery and logically justified the subsequent diagnosis of acute posthemorrhagic anemia (D62). 
LLM auditors (Gemini, Qwen) similarly highlighted the highly realistic temporal trends in coagulation monitoring and the use of guideline-concordant discharge medications (e.g., aspirin, statins, beta-blockers) as key factors driving the high realism score.}
\label{tab:case-study-022}\\
\toprule
\textbf{time} & \textbf{code} & \textbf{numerical\_value} & \textbf{code\_label} \\
\midrule
\endfirsthead

\multicolumn{4}{c}{\tablename~\thetable{} -- \textit{continued from previous page}}\\
\midrule
\textbf{time} & \textbf{code} & \textbf{numerical\_value} & \textbf{code\_label} \\
\midrule
\endhead

\midrule
\multicolumn{4}{r}{\textit{continued on next page}}\\
\endfoot

\bottomrule
\endlastfoot

11-29 22:01 & \texttt{START\_RECORD} & & \\
11-29 22:01 & \texttt{AGE\_75\_80\_years} & & \\
11-29 22:01 & \texttt{SEX\_F} & & \\
11-29 22:01 & \texttt{RACE\_UNKNOWN} & & \\
11-29 22:01 & \texttt{MARITAL\_STATUS\_MARRIED} & & \\
11-29 22:01 & \texttt{YEAR\_2014} & & \\ 
11-29 22:01 & \texttt{START\_VISIT} & & \\
11-29 22:01 & \texttt{LAB\_Hematocrit\_\%} & 35$\cdot$51 & \\
11-29 22:01 & \texttt{LAB\_Hemoglobin\_g/dL} & 10$\cdot$99 & \\
11-29 22:01 & \texttt{LAB\_MCHC\_\%} & 32$\cdot$08 & \\
11-29 22:01 & \texttt{LAB\_MCH\_pg} & 31$\cdot$14 & \\
11-29 22:01 & \texttt{LAB\_MCV\_fL} & 98 & \\
11-29 22:01 & \texttt{LAB\_Platelet\_Count\_K/uL} & 248 & \\
11-29 22:01 & \texttt{LAB\_RDW\_\%} & 12$\cdot$6 & \\
11-29 22:01 & \texttt{LAB\_Red\_Blood\_Cells\_m/uL} & 3$\cdot$43 & \\
11-29 22:01 & \texttt{LAB\_White\_Blood\_Cells\_K/uL} & 6$\cdot$26 & \\
11-29 22:08 & \texttt{LAB\_INR(PT)\_NO\_UNIT} & 1$\cdot$06 & \\
11-29 22:08 & \texttt{LAB\_PT\_sec} & 11$\cdot$82 & \\
11-29 22:17 & \texttt{LAB\_PTT\_sec} & 33$\cdot$56 & \\
11-29 22:27 & \texttt{LAB\_Anion\_Gap\_mEq/L} & 12 & \\
11-29 22:27 & \texttt{LAB\_Bicarbonate\_mEq/L} & 27 & \\
11-29 22:27 & \texttt{LAB\_Calcium\_\_Total\_mg/dL} & 8$\cdot$66 & \\
11-29 22:27 & \texttt{LAB\_Chloride\_mEq/L} & 102 & \\
11-29 22:27 & \texttt{LAB\_Creatine\_Kinase\_\_MB\_Isoenzyme\_ng/mL} & 2 & \\
11-29 22:27 & \texttt{LAB\_Creatinine\_mg/dL} & 1$\cdot$14 & \\
11-29 22:27 & \texttt{LAB\_Magnesium\_mg/dL} & 2$\cdot$2 & \\
11-29 22:27 & \texttt{LAB\_Phosphate\_mg/dL} & 3$\cdot$21 & \\
11-29 22:27 & \texttt{LAB\_Potassium\_mEq/L} & 3$\cdot$25 & \\
11-29 22:27 & \texttt{LAB\_Sodium\_mEq/L} & 137 & \\
11-29 22:27 & \texttt{LAB\_Urea\_Nitrogen\_mg/dL} & 12 & \\
11-30 10:24 & \texttt{LAB\_Cholesterol\_Ratio\_(Total/HDL)\_Ratio} & 1$\cdot$76 & \\
11-30 10:24 & \texttt{LAB\_Cholesterol\_\_HDL\_mg/dL} & 61 & \\
11-30 10:24 & \texttt{LAB\_Cholesterol\_\_LDL\_\_Calculated\_mg/dL} & 57 & \\
11-30 10:24 & \texttt{LAB\_Cholesterol\_\_Total\_mg/dL} & 126 & \\
11-30 10:24 & \texttt{LAB\_Thyroid\_Stimulating\_Hormone\_uIU/mL} & 4$\cdot$81 & \\
11-30 18:57 & \texttt{LAB\_PTT\_sec} & 64$\cdot$99 & \\
11-30 23:14 & \texttt{LAB\_Hematocrit\_\%} & 34$\cdot$54 & \\
11-30 23:14 & \texttt{LAB\_Hemoglobin\_g/dL} & 11$\cdot$02 & \\
11-30 23:14 & \texttt{LAB\_MCHC\_\%} & 31$\cdot$41 & \\
11-30 23:14 & \texttt{LAB\_MCH\_pg} & 31$\cdot$7 & \\
11-30 23:14 & \texttt{LAB\_MCV\_fL} & 98 & \\
11-30 23:14 & \texttt{LAB\_Platelet\_Count\_K/uL} & 268 & \\
11-30 23:14 & \texttt{LAB\_RDW\_\%} & 12$\cdot$24 & \\
11-30 23:14 & \texttt{LAB\_Red\_Blood\_Cells\_m/uL} & 3$\cdot$41 & \\
11-30 23:14 & \texttt{LAB\_White\_Blood\_Cells\_K/uL} & 6$\cdot$66 & \\
11-30 23:57 & \texttt{LAB\_Anion\_Gap\_mEq/L} & 12 & \\
11-30 23:57 & \texttt{LAB\_Bicarbonate\_mEq/L} & 27 & \\
11-30 23:57 & \texttt{LAB\_Calcium\_\_Total\_mg/dL} & 8$\cdot$98 & \\
11-30 23:57 & \texttt{LAB\_Chloride\_mEq/L} & 104 & \\
11-30 23:57 & \texttt{LAB\_Creatinine\_mg/dL} & 1$\cdot$33 & \\
11-30 23:57 & \texttt{LAB\_Magnesium\_mg/dL} & 2$\cdot$37 & \\
11-30 23:57 & \texttt{LAB\_Phosphate\_mg/dL} & 3$\cdot$7 & \\
11-30 23:57 & \texttt{LAB\_Potassium\_mEq/L} & 3$\cdot$7 & \\
11-30 23:57 & \texttt{LAB\_Sodium\_mEq/L} & 143 & \\
11-30 23:57 & \texttt{LAB\_Urea\_Nitrogen\_mg/dL} & 12 & \\
12-01 00:27 & \texttt{LAB\_INR(PT)\_NO\_UNIT} & 1$\cdot$09 & \\
12-01 00:27 & \texttt{LAB\_PTT\_sec} & 57$\cdot$26 & \\
12-01 00:27 & \texttt{LAB\_PT\_sec} & 11$\cdot$79 & \\
12-01 02:14 & \texttt{LAB\_PTT\_sec} & 57$\cdot$31 & \\
12-01 10:54 & \texttt{LAB\_PTT\_sec} & 62$\cdot$03 & \\
12-01 16:43 & \texttt{LAB\_PTT\_sec} & 62$\cdot$94 & \\
12-02 04:40 & \texttt{LAB\_Hematocrit\_\%} & 34$\cdot$03 & \\
12-02 04:40 & \texttt{LAB\_Hemoglobin\_g/dL} & 11$\cdot$42 & \\
12-02 04:40 & \texttt{LAB\_MCHC\_\%} & 32$\cdot$04 & \\
12-02 04:40 & \texttt{LAB\_MCH\_pg} & 31$\cdot$61 & \\
12-02 04:40 & \texttt{LAB\_MCV\_fL} & 96 & \\
12-02 04:40 & \texttt{LAB\_Platelet\_Count\_K/uL} & 300 & \\
12-02 04:40 & \texttt{LAB\_RDW\_\%} & 12$\cdot$73 & \\
12-02 04:40 & \texttt{LAB\_Red\_Blood\_Cells\_m/uL} & 3$\cdot$47 & \\
12-02 04:40 & \texttt{LAB\_White\_Blood\_Cells\_K/uL} & 5$\cdot$54 & \\
12-02 05:08 & \texttt{LAB\_INR(PT)\_NO\_UNIT} & 1$\cdot$07 & \\
12-02 05:08 & \texttt{LAB\_PTT\_sec} & 62$\cdot$07 & \\
12-02 05:08 & \texttt{LAB\_PT\_sec} & 11$\cdot$56 & \\
12-02 05:55 & \texttt{LAB\_Anion\_Gap\_mEq/L} & 14 & \\
12-02 05:55 & \texttt{LAB\_Bicarbonate\_mEq/L} & 29 & \\
12-02 05:55 & \texttt{LAB\_Calcium\_\_Total\_mg/dL} & 8$\cdot$72 & \\
12-02 05:55 & \texttt{LAB\_Chloride\_mEq/L} & 103 & \\
12-02 05:55 & \texttt{LAB\_Creatinine\_mg/dL} & 1$\cdot$32 & \\
12-02 05:55 & \texttt{LAB\_Magnesium\_mg/dL} & 2$\cdot$17 & \\
12-02 05:55 & \texttt{LAB\_Phosphate\_mg/dL} & 4$\cdot$04 & \\
12-02 05:55 & \texttt{LAB\_Potassium\_mEq/L} & 3$\cdot$1 & \\
12-02 05:55 & \texttt{LAB\_Sodium\_mEq/L} & 143 & \\
12-02 05:55 & \texttt{LAB\_Urea\_Nitrogen\_mg/dL} & 13 & \\
12-02 13:22 & \texttt{LAB\_PTT\_sec} & 72$\cdot$54 & \\
12-02 18:41 & \texttt{ICD10PCS\_021048C} & & Bypass Coronary Artery, One Artery from Thoracic Artery with Zooplastic Tissue, Percutaneous Endoscopic Approach \\
12-02 18:41 & \texttt{ICD10PCS\_02134KW} & & Bypass Coronary Artery, Four or More Arteries from Aorta with Nonautologous Tissue Substitute, Percutaneous Endoscopic Approach \\
12-02 18:41 & \texttt{ICD10PCS\_5A1221Z} & & Performance of Cardiac Output, Continuous \\
12-03 04:15 & \texttt{LAB\_Base\_Excess\_mEq/L} & 0 & \\
12-03 04:15 & \texttt{LAB\_Calculated\_Total\_CO2\_mEq/L} & 27 & \\
12-03 04:15 & \texttt{LAB\_Chloride\_\_Whole\_Blood\_mEq/L} & 104 & \\
12-03 04:15 & \texttt{LAB\_Free\_Calcium\_mmol/L} & 1$\cdot$16 & \\
12-03 04:15 & \texttt{LAB\_Glucose\_mg/dL} & 106 & \\
12-03 04:15 & \texttt{LAB\_Hematocrit\_\_Calculated\_\%} & 33 & \\
12-03 04:15 & \texttt{LAB\_Hemoglobin\_g/dL} & 11$\cdot$03 & \\
12-03 04:15 & \texttt{LAB\_Lactate\_mmol/L} & 0$\cdot$64 & \\
12-03 04:15 & \texttt{LAB\_Potassium\_\_Whole\_Blood\_mEq/L} & 3$\cdot$29 & \\
12-03 04:15 & \texttt{LAB\_Sodium\_\_Whole\_Blood\_mEq/L} & 136 & \\
12-03 04:15 & \texttt{LAB\_pCO2\_mm\_Hg} & 51 & \\
12-03 04:15 & \texttt{LAB\_pH\_units} & 7$\cdot$32 & \\
12-03 04:15 & \texttt{LAB\_pO2\_mm\_Hg} & 365 & \\
12-03 09:58 & \texttt{LAB\_Base\_Excess\_mEq/L} & 1 & \\
12-03 09:58 & \texttt{LAB\_Calculated\_Total\_CO2\_mEq/L} & 27 & \\
12-03 09:58 & \texttt{LAB\_Glucose\_mg/dL} & 138 & \\
12-03 09:58 & \texttt{LAB\_Hematocrit\_\_Calculated\_\%} & 20 & \\
12-03 09:58 & \texttt{LAB\_Hemoglobin\_g/dL} & 7$\cdot$36 & \\
12-03 09:58 & \texttt{LAB\_Potassium\_\_Whole\_Blood\_mEq/L} & 4$\cdot$43 & \\
12-03 09:58 & \texttt{LAB\_pCO2\_mm\_Hg} & 42 & \\
12-03 09:58 & \texttt{LAB\_pH\_units} & 7$\cdot$41 & \\
12-03 09:58 & \texttt{LAB\_pO2\_mm\_Hg} & 396 & \\
12-03 10:48 & \texttt{LAB\_Base\_Excess\_mEq/L} & 5 & \\
12-03 10:48 & \texttt{LAB\_Calculated\_Total\_CO2\_mEq/L} & 26 & \\
12-03 10:48 & \texttt{LAB\_Glucose\_mg/dL} & 174 & \\
12-03 10:48 & \texttt{LAB\_Hematocrit\_\_Calculated\_\%} & 21 & \\
12-03 10:48 & \texttt{LAB\_Hemoglobin\_g/dL} & 7$\cdot$52 & \\
12-03 10:48 & \texttt{LAB\_Potassium\_\_Whole\_Blood\_mEq/L} & 4$\cdot$82 & \\
12-03 10:48 & \texttt{LAB\_pCO2\_mm\_Hg} & 32 & \\
12-03 10:48 & \texttt{LAB\_pH\_units} & 7$\cdot$89 & \\
12-03 10:48 & \texttt{LAB\_pO2\_mm\_Hg} & 310 & \\
12-03 11:30 & \texttt{LAB\_Base\_Excess\_mEq/L} & 2 & \\
12-03 11:30 & \texttt{LAB\_Calculated\_Total\_CO2\_mEq/L} & 25 & \\
12-03 11:30 & \texttt{LAB\_Chloride\_\_Whole\_Blood\_mEq/L} & 106 & \\
12-03 11:30 & \texttt{LAB\_Free\_Calcium\_mmol/L} & 1$\cdot$21 & \\
12-03 11:30 & \texttt{LAB\_Glucose\_mg/dL} & 165 & \\
12-03 11:30 & \texttt{LAB\_Hematocrit\_\_Calculated\_\%} & 25 & \\
12-03 11:30 & \texttt{LAB\_Hemoglobin\_g/dL} & 8$\cdot$13 & \\
12-03 11:30 & \texttt{LAB\_Lactate\_mmol/L} & 2$\cdot$34 & \\
12-03 11:30 & \texttt{LAB\_Potassium\_\_Whole\_Blood\_mEq/L} & 3$\cdot$97 & \\
12-03 11:30 & \texttt{LAB\_Sodium\_\_Whole\_Blood\_mEq/L} & 132 & \\
12-03 11:30 & \texttt{LAB\_pCO2\_mm\_Hg} & 34 & \\
12-03 11:30 & \texttt{LAB\_pH\_units} & 7$\cdot$47 & \\
12-03 11:30 & \texttt{LAB\_pO2\_mm\_Hg} & 297 & \\
12-03 11:30 & \texttt{LAB\_Platelet\_Count\_K/uL} & 226 & \\
12-03 11:37 & \texttt{LAB\_Fibrinogen\_\_Functional\_mg/dL} & 251 & \\
12-03 11:37 & \texttt{LAB\_INR(PT)\_NO\_UNIT} & 1$\cdot$65 & \\
12-03 11:37 & \texttt{LAB\_PT\_sec} & 17$\cdot$06 & \\
12-03 11:50 & \texttt{LAB\_PTT\_sec} & 27$\cdot$5 & \\
12-03 12:22 & \texttt{LAB\_Base\_Excess\_mEq/L} & 2 & \\
12-03 12:22 & \texttt{LAB\_Calculated\_Total\_CO2\_mEq/L} & 26 & \\
12-03 12:22 & \texttt{LAB\_pCO2\_mm\_Hg} & 36 & \\
12-03 12:22 & \texttt{LAB\_pH\_units} & 7$\cdot$47 & \\
12-03 12:22 & \texttt{LAB\_pO2\_mm\_Hg} & 311 & \\
12-03 12:22 & \texttt{LAB\_Free\_Calcium\_mmol/L} & 1$\cdot$3 & \\
12-03 12:22 & \texttt{LAB\_Glucose\_mg/dL} & 135 & \\
12-03 12:22 & \texttt{LAB\_Potassium\_\_Whole\_Blood\_mEq/L} & 3$\cdot$97 & \\
12-03 12:22 & \texttt{LAB\_Sodium\_\_Whole\_Blood\_mEq/L} & 134 & \\
12-03 12:29 & \texttt{LAB\_Hematocrit\_\%} & 32$\cdot$88 & \\
12-03 12:29 & \texttt{LAB\_Hemoglobin\_g/dL} & 10$\cdot$57 & \\
12-03 12:29 & \texttt{LAB\_MCHC\_\%} & 32$\cdot$25 & \\
12-03 12:29 & \texttt{LAB\_MCH\_pg} & 31$\cdot$67 & \\
12-03 12:29 & \texttt{LAB\_MCV\_fL} & 97 & \\
12-03 12:29 & \texttt{LAB\_Platelet\_Count\_K/uL} & 240 & \\
12-03 12:29 & \texttt{LAB\_RDW\_\%} & 12$\cdot$54 & \\
12-03 12:29 & \texttt{LAB\_Red\_Blood\_Cells\_m/uL} & 3$\cdot$36 & \\
12-03 12:29 & \texttt{LAB\_White\_Blood\_Cells\_K/uL} & 13$\cdot$7 & \\
12-03 12:36 & \texttt{LAB\_INR(PT)\_NO\_UNIT} & 1$\cdot$14 & \\
12-03 12:36 & \texttt{LAB\_PTT\_sec} & 27$\cdot$84 & \\
12-03 12:36 & \texttt{LAB\_PT\_sec} & 13$\cdot$17 & \\
12-03 13:04 & \texttt{LAB\_Anion\_Gap\_mEq/L} & 14 & \\
12-03 13:04 & \texttt{LAB\_Bicarbonate\_mEq/L} & 23 & \\
12-03 13:04 & \texttt{LAB\_Chloride\_mEq/L} & 104 & \\
12-03 13:04 & \texttt{LAB\_Creatinine\_mg/dL} & 0$\cdot$78 & \\
12-03 13:04 & \texttt{LAB\_Potassium\_mEq/L} & 3$\cdot$89 & \\
12-03 13:04 & \texttt{LAB\_Sodium\_mEq/L} & 137 & \\
12-03 13:04 & \texttt{LAB\_Urea\_Nitrogen\_mg/dL} & 9 & \\
12-03 17:24 & \texttt{LAB\_Base\_Excess\_mEq/L} & 0 & \\
12-03 17:24 & \texttt{LAB\_Calculated\_Total\_CO2\_mEq/L} & 24 & \\
12-03 17:24 & \texttt{LAB\_Free\_Calcium\_mmol/L} & 1$\cdot$14 & \\
12-03 17:24 & \texttt{LAB\_Glucose\_mg/dL} & 117 & \\
12-03 17:24 & \texttt{LAB\_Oxygen\_Saturation\_\%} & 98 & \\
12-03 17:24 & \texttt{LAB\_Potassium\_\_Whole\_Blood\_mEq/L} & 3$\cdot$68 & \\
12-03 17:24 & \texttt{LAB\_pCO2\_mm\_Hg} & 37 & \\
12-03 17:24 & \texttt{LAB\_pH\_units} & 7$\cdot$43 & \\
12-03 17:24 & \texttt{LAB\_pO2\_mm\_Hg} & 173 & \\
12-03 22:35 & \texttt{LAB\_Chloride\_mEq/L} & 111 & \\
12-03 22:35 & \texttt{LAB\_Potassium\_mEq/L} & 4$\cdot$05 & \\
12-03 22:35 & \texttt{LAB\_Sodium\_mEq/L} & 139 & \\
12-03 22:35 & \texttt{LAB\_Hematocrit\_\%} & 32$\cdot$33 & \\
12-03 23:36 & \texttt{ICD10PCS\_0JHG0XZ} & & Insertion of Tunneled Vascular Access Device into Right Lower Arm Subcutaneous Tissue and Fascia, Open Approach \\
12-04 01:40 & \texttt{LAB\_Hematocrit\_\%} & 28$\cdot$6 & \\
12-04 01:40 & \texttt{LAB\_Hemoglobin\_g/dL} & 9$\cdot$16 & \\
12-04 01:40 & \texttt{LAB\_MCHC\_\%} & 32$\cdot$79 & \\
12-04 01:40 & \texttt{LAB\_MCH\_pg} & 32$\cdot$63 & \\
12-04 01:40 & \texttt{LAB\_MCV\_fL} & 97 & \\
12-04 01:40 & \texttt{LAB\_Platelet\_Count\_K/uL} & 221 & \\
12-04 01:40 & \texttt{LAB\_RDW\_\%} & 12$\cdot$76 & \\
12-04 01:40 & \texttt{LAB\_Red\_Blood\_Cells\_m/uL} & 2$\cdot$8 & \\
12-04 01:40 & \texttt{LAB\_White\_Blood\_Cells\_K/uL} & 10$\cdot$8 & \\
12-04 02:24 & \texttt{LAB\_Anion\_Gap\_mEq/L} & 10 & \\
12-04 02:24 & \texttt{LAB\_Bicarbonate\_mEq/L} & 26 & \\
12-04 02:24 & \texttt{LAB\_Chloride\_mEq/L} & 109 & \\
12-04 02:24 & \texttt{LAB\_Creatinine\_mg/dL} & 1$\cdot$02 & \\
12-04 02:24 & \texttt{LAB\_Potassium\_mEq/L} & 4$\cdot$29 & \\
12-04 02:24 & \texttt{LAB\_Sodium\_mEq/L} & 138 & \\
12-04 02:24 & \texttt{LAB\_Urea\_Nitrogen\_mg/dL} & 9 & \\
12-07 00:35 & \texttt{LAB\_Hematocrit\_\%} & 24$\cdot$34 & \\
12-07 00:35 & \texttt{LAB\_Hemoglobin\_g/dL} & 7$\cdot$91 & \\
12-07 00:35 & \texttt{LAB\_MCHC\_\%} & 32$\cdot$04 & \\
12-07 00:35 & \texttt{LAB\_MCH\_pg} & 31$\cdot$86 & \\
12-07 00:35 & \texttt{LAB\_MCV\_fL} & 99 & \\
12-07 00:35 & \texttt{LAB\_Platelet\_Count\_K/uL} & 139 & \\
12-07 00:35 & \texttt{LAB\_RDW\_\%} & 12$\cdot$7 & \\
12-07 00:35 & \texttt{LAB\_Red\_Blood\_Cells\_m/uL} & 2$\cdot$73 & \\
12-07 00:35 & \texttt{LAB\_White\_Blood\_Cells\_K/uL} & 9$\cdot$0 & \\
12-07 01:26 & \texttt{LAB\_Anion\_Gap\_mEq/L} & 13 & \\
12-07 01:26 & \texttt{LAB\_Bicarbonate\_mEq/L} & 23 & \\
12-07 01:26 & \texttt{LAB\_Chloride\_mEq/L} & 105 & \\
12-07 01:26 & \texttt{LAB\_Creatinine\_mg/dL} & 0$\cdot$89 & \\
12-07 01:26 & \texttt{LAB\_Potassium\_mEq/L} & 4$\cdot$06 & \\
12-07 01:26 & \texttt{LAB\_Sodium\_mEq/L} & 137 & \\
12-07 01:26 & \texttt{LAB\_Urea\_Nitrogen\_mg/dL} & 9 & \\
12-07 12:51 & \texttt{LAB\_Hematocrit\_\%} & 27$\cdot$65 & \\
12-07 15:51 & \texttt{ATC\_N02BE01} & & paracetamol (acetaminophen) \\
12-07 16:08 & \texttt{ATC\_A10AE05} & & insulin detemir \\
12-07 16:08 & \texttt{ATC\_A12CA01} & & sodium chloride \\
12-07 16:08 & \texttt{ATC\_J01DB04} & & cefazolin \\
12-07 22:07 & \texttt{ATC\_A02AA04} & & magnesium hydroxide \\
12-07 22:07 & \texttt{ATC\_A02BA02} & & ranitidine \\
12-07 22:07 & \texttt{ATC\_A06AA} & & docusate \\
12-07 22:07 & \texttt{ATC\_A12BA01} & & potassium chloride \\
12-07 22:07 & \texttt{ATC\_C07AB02} & & metoprolol \\
12-07 22:07 & \texttt{ATC\_C10AA05} & & atorvastatin \\
12-07 23:55 & \texttt{ATC\_A10AE05} & & insulin detemir \\
12-08 01:42 & \texttt{ATC\_N02BE01} & & paracetamol (acetaminophen) \\
12-08 02:43 & \texttt{ATC\_J01DB04} & & cefazolin \\
12-08 04:14 & \texttt{ATC\_N02BE01} & & paracetamol (acetaminophen) \\
12-08 08:21 & \texttt{LAB\_Hematocrit\_\%} & 26$\cdot$85 & \\
12-08 08:21 & \texttt{LAB\_Hemoglobin\_g/dL} & 8$\cdot$72 & \\
12-08 08:21 & \texttt{LAB\_MCHC\_\%} & 32$\cdot$83 & \\
12-08 08:21 & \texttt{LAB\_MCH\_pg} & 32$\cdot$32 & \\
12-08 08:21 & \texttt{LAB\_MCV\_fL} & 99 & \\
12-08 08:21 & \texttt{LAB\_Platelet\_Count\_K/uL} & 207 & \\
12-08 08:21 & \texttt{LAB\_RDW\_\%} & 12$\cdot$93 & \\
12-08 08:21 & \texttt{LAB\_Red\_Blood\_Cells\_m/uL} & 2$\cdot$76 & \\
12-08 08:21 & \texttt{LAB\_White\_Blood\_Cells\_K/uL} & 8$\cdot$61 & \\
12-08 08:55 & \texttt{LAB\_Anion\_Gap\_mEq/L} & 7 & \\
12-08 08:55 & \texttt{LAB\_Bicarbonate\_mEq/L} & 28 & \\
12-08 08:55 & \texttt{LAB\_Calcium\_\_Total\_mg/dL} & 8$\cdot$18 & \\
12-08 08:55 & \texttt{LAB\_Chloride\_mEq/L} & 105 & \\
12-08 08:55 & \texttt{LAB\_Creatinine\_mg/dL} & 1$\cdot$05 & \\
12-08 08:55 & \texttt{LAB\_Magnesium\_mg/dL} & 2$\cdot$3 & \\
12-08 08:55 & \texttt{LAB\_Phosphate\_mg/dL} & 2$\cdot$5 & \\
12-08 08:55 & \texttt{LAB\_Potassium\_mEq/L} & 3$\cdot$65 & \\
12-08 08:55 & \texttt{LAB\_Sodium\_mEq/L} & 138 & \\
12-08 08:55 & \texttt{LAB\_Urea\_Nitrogen\_mg/dL} & 16 & \\
12-08 09:17 & \texttt{ATC\_A01AD05} & & acetylsalicylic acid \\
12-08 09:17 & \texttt{ATC\_A02BA02} & & ranitidine \\
12-08 09:17 & \texttt{ATC\_A06AA} & & docusate \\
12-08 09:17 & \texttt{ATC\_A10AE05} & & insulin detemir \\
12-08 09:17 & \texttt{ATC\_A12BA01} & & potassium chloride \\
12-08 09:17 & \texttt{ATC\_C03CA01} & & furosemide \\
12-08 09:17 & \texttt{ATC\_C07AB02} & & metoprolol \\
12-08 12:28 & \texttt{ATC\_J07AL} & & pneumococcal vaccines \\
12-08 12:48 & \texttt{LAB\_Hematocrit\_\%} & 28$\cdot$03 & \\
12-08 12:48 & \texttt{LAB\_Hemoglobin\_g/dL} & 8$\cdot$91 & \\
12-08 12:48 & \texttt{LAB\_MCHC\_\%} & 32$\cdot$02 & \\
12-08 12:48 & \texttt{LAB\_MCH\_pg} & 31$\cdot$11 & \\
12-08 12:48 & \texttt{LAB\_MCV\_fL} & 97 & \\
12-08 12:48 & \texttt{LAB\_Platelet\_Count\_K/uL} & 193 & \\
12-08 12:48 & \texttt{LAB\_RDW\_\%} & 13$\cdot$16 & \\
12-08 12:48 & \texttt{LAB\_Red\_Blood\_Cells\_m/uL} & 2$\cdot$84 & \\
12-08 12:48 & \texttt{LAB\_White\_Blood\_Cells\_K/uL} & 7$\cdot$86 & \\
12-08 12:56 & \texttt{ATC\_A12BA01} & & potassium chloride \\
12-08 12:56 & \texttt{ATC\_N02BE01} & & paracetamol (acetaminophen) \\
12-08 17:52 & \texttt{ATC\_A10AE05} & & insulin detemir \\
12-08 21:43 & \texttt{ICD10CM\_D62} & & Acute posthemorrhagic anemia \\
12-08 21:43 & \texttt{ICD10CM\_E119} & & Type 2 diabetes mellitus without complications \\
12-08 21:43 & \texttt{ICD10CM\_E784} & & Other hyperlipidemia \\
12-08 21:43 & \texttt{ICD10CM\_F329} & & Major depressive disorder, single episode, unspecified \\
12-08 21:43 & \texttt{ICD10CM\_I10} & & Essential (primary) hypertension \\
12-08 21:43 & \texttt{ICD10CM\_I208} & & Other forms of angina pectoris \\
12-08 21:43 & \texttt{ICD10CM\_I2510} & & Atherosclerotic heart disease of native coronary artery without angina pectoris \\
12-08 21:43 & \texttt{ICD10CM\_K219} & & Gastro-esophageal reflux disease without esophagitis \\
12-08 21:43 & \texttt{ICD10CM\_M129} & & Arthropathy, unspecified \\
12-08 21:43 & \texttt{ICD10CM\_Z853} & & Personal history of malignant neoplasm of breast \\
12-08 21:43 & \texttt{ICD10CM\_Z87891} & & Personal history of nicotine dependence \\
12-08 21:43 & \texttt{ICD10CM\_Z923} & & Personal history of irradiation \\
12-08 21:43 & \texttt{END\_VISIT} & & \\
12-08 21:43 & \texttt{END\_RECORD} & & \\
\end{longtable}



\end{document}